\documentclass[10pt,journal,compsoc]{IEEEtran}

\usepackage{nicefrac}       % compact symbols for 1/2, etc.
\usepackage{epsfig}
\usepackage{graphicx}
\usepackage{amsmath}
\usepackage{amssymb}
\usepackage{url}
\usepackage{amsthm}
\usepackage{algorithm}               %format of the algorithm
\usepackage{algorithmic}             %format of the algorithm
\usepackage{multirow}                %multirow for format of  table
\usepackage{bm}
\usepackage{footmisc}
\usepackage{color}
\usepackage{booktabs}       % professional-quality tables
\usepackage{color}

\ifCLASSOPTIONcompsoc
  \usepackage[nocompress]{cite}
\else
  \usepackage{cite}
\fi
\ifCLASSINFOpdf
\else
\fi
\allowdisplaybreaks

\def\A{\tilde{A}}

\def\E{\tilde{E}}
\def\R{\mathbb{R}}

\title{Fourier Series Expansion Based Filter Parametrization for Equivariant Convolutions}

\author{%
  David S.~Hippocampus\thanks{Use footnote for providing further information
    about author (webpage, alternative address)---\emph{not} for acknowledging
    funding agencies.} \\
  Department of Computer Science\\
  Cranberry-Lemon University\\
  Pittsburgh, PA 15213 \\
  \texttt{hippo@cs.cranberry-lemon.edu} \\
}
\begin{document}

\title{Fourier Series Expansion Based Filter Parametrization for Equivariant Convolutions}
\author{Qi Xie,   Qian Zhao, Zongben Xu and Deyu Meng
\thanks{ Qi Xie, Qian Zhao, Zongben Xu and Deyu Meng (corresponding author) are with School of Mathematics and Statistics and Ministry of Education Key Lab of Intelligent Networks and Network Security, Xi'an Jiaotong University, Shaanxi, P.R.China., and Peng Cheng Laboratory, Shenzhen, P.R. China.}
\thanks{Email: \{xie.qi, timmy.zhaoqian, zbxu, dymeng\}@mail.xjtu.edu.cn}
%\thanks{Manuscript received April 19, 2019; revised August 26, 2015.}
	%$^\ast${\small Corresponding author}
}

\markboth{IEEE Transactions on Pattern Analysis and Machine Intelligence, 2021}
{Xie \MakeLowercase{\textit{et al.}}: Fourier Series Expansion Based Filter Parametrization for Equivariant Convolutions}

\IEEEtitleabstractindextext{
\begin{abstract}
It has been shown that equivariant convolution is very helpful for many types of computer vision tasks. Recently, the 2D filter parametrization technique  {has played an important role for designing} equivariant convolutions, and has achieved success in making use of rotation symmetry of images.
However, the current filter parametrization strategy still has its evident drawbacks, where the most critical one lies in the accuracy problem of filter representation.  {To address this issue, in this paper we explore an ameliorated Fourier series expansion for 2D filters,
and propose a new filter parametrization method based on it.}
The proposed filter parametrization method not only  finely represents 2D filters with zero error when the filter is not rotated (similar as the classical Fourier series expansion),
but also substantially alleviates the  {aliasing-effect-caused} quality degradation when the filter is rotated (which usually arises in classical Fourier series expansion method).
Accordingly, we construct a new equivariant convolution method based on the proposed filter parametrization method, named F-Conv. We prove that the equivariance of the proposed F-Conv is exact in the continuous domain, which becomes approximate only after discretization.
Moreover, we provide theoretical error analysis for the case when the equivariance is approximate, showing that the approximation error is related to the mesh size and filter size. Extensive experiments show the superiority of the proposed method. Particularly, we adopt rotation equivariant convolution methods to a typical low-level image processing task, image super-resolution. It can be substantiated that the proposed F-Conv based method evidently outperforms classical convolution based methods.
 {Compared with pervious filter parametrization based methods, the F-Conv performs more accurately on this low-level image processing task, reflecting its intrinsic capability of faithfully preserving rotation symmetries in local image features.}
\end{abstract}

}
%\mathfrak{}
\maketitle
\IEEEdisplaynontitleabstractindextext
\IEEEpeerreviewmaketitle

\newtheorem{Thm}{Theorem}
\newtheorem{Rem}{Remark}

\IEEEraisesectionheading{\section{Introduction}\label{sec:introduction}}

\IEEEPARstart{I}n many computer vision and image processing tasks, there are usually kinds of transformation symmetries existed across both the local features and global semantic representations of images \cite{zeiler2014visualizing}.
 {Typical examples include translation, rotation and reflection symmetries, intrinsically contained by different kinds of images utilized in  both low-level and high-level computer vision tasks.}
It is desirable to make use of such transformation symmetries to reduce the number of model parameters and enhance the generalization capability of machine learning methods.

In the past few years, convolutional neural network (CNN) based models have achieved great success in many computer vision and image processing tasks,
such as image recognition, objective detection, semantic segmentation and image reconstruction \cite{szegedy2015going}.
One of the most principal reasons is that CNN is a shift equivariant network. That is, shifting an input image of CNN is equivalent to shifting all of its intermediate feature maps and output image.
In other words, the translation symmetry is preserved  {throughout the CNN layers}. Compared with conventional fully-connected neural networks,  this translational equivariance property brings in weight sharing for CNN, makes the network parameters being used more efficiently, and thus leads to substantially better generalization capability.
Recently, several kinds of equivariant CNNs have been further proposed to preserve rotations and reflection symmetries beyond current CNNs.
It has been shown that equivariant CNNs can be very helpful for computer vision tasks on images like biomedical microscopy and astronomical images
\cite{weiler2018learning, weiler2019general, shen2020pdo}.

The early equivariant CNNs achieve $\nicefrac{\pi}{2}$ degree rotations and reflection equivariant (i.e., $p4$ and $p4m$ group equivariant) on square lattices \cite{Cohen2016group}. This kind of CNNs is designed based on the property that $\nicefrac{\pi}{2}$  degree rotation and reflection on square lattice can be easily implemented by changing the position of elements in a convolutional filter, and an element-shared filter set can be easily designed in this way.
Afterwards, HexaConv \cite{hoogeboom2018hexaconv} expanded rotation equivariant to  $\nicefrac{\pi}{3}$  degree rotations (i.e., $p6$ and $p6m$ group equivariant) by replacing the commonly used square lattices with hexagonal lattices.
However, this category of  methods can only deal with a 4-fold or 6-fold rotational symmetry for images, since the practically used image data are mostly stored with square lattices and it is hard to be transformed into lattices other than hexagonal one.

\begin{figure*}[t]\vspace{1mm}
\begin{center}
\hspace{-0mm}\includegraphics[width=1\linewidth]{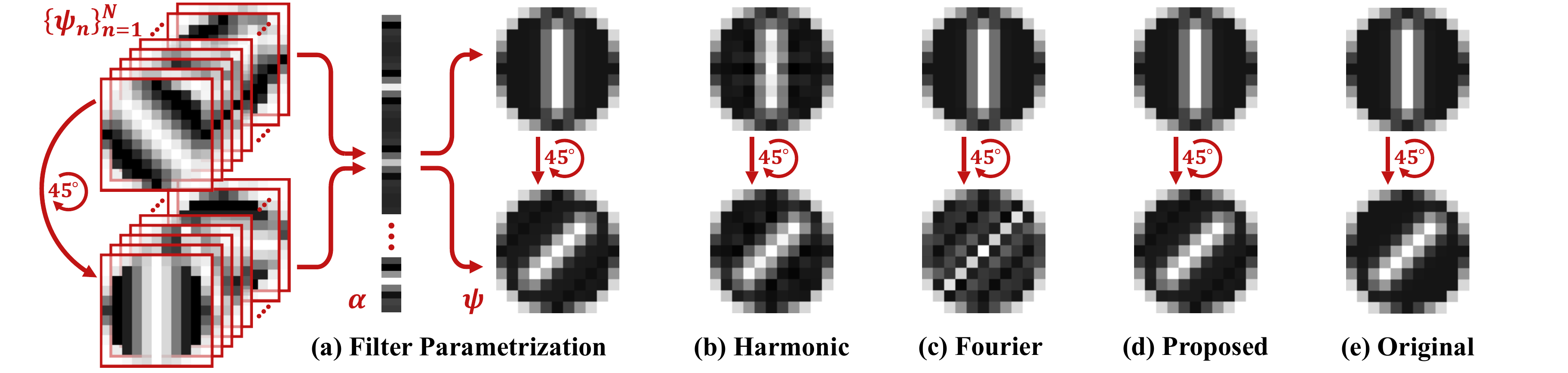}
\end{center}
\vspace{-5mm}
  \caption{(a) Illustration of filter parametrization by linear combination of basis functions,  and the filter rotation based on it. All the filters are with circular shape masks for better rotation.
   (b)-(d) The representations (upper) and correlated $\nicefrac{\pi}{4}$ rotations (lower) of a given 2D filter, by adopting harmonics bases \cite{weiler2018learning},
   2D Fourier bases \cite{pei1998two} and the proposed bases as the basis functions, respectively. (e) The original filter (upper) and its $\nicefrac{\pi}{4}$ rotation (lower).}
\label{Fig:Filter}
\vspace{-1.5mm}
\end{figure*}

Very recently, the filter parametrization technique has been employed to make use of more rotation symmetry in CNNs \cite{weiler2018learning, weiler2019general, shen2020pdo}.
In 2018, Weiler et al. \cite{weiler2018learning} proposed harmonics based steerable filters to achieve equivariance on arbitrary degree rotation in the continuous domain. Later on, Shen et al. \cite{shen2020pdo} further proposed partial differential operator based equivariant convolution (PDO-eConv).
The basic idea of these methods is to define the to-be-learnt filters as
the linear combination of a set of basis functions
(i.e., elementary filters) and learn the combination coefficients \cite{freeman1991design}.
Then, as the example shown in Fig. \ref{Fig:Filter} (a),
one can rotate the filter to arbitrary degrees by rotating the basis functions in the continuous domain and use it to construct rotation equivariant convolutions.

 {While the current filter parametrization strategies have achieved great success in the development of equivariant CNNs, they still have evident drawbacks.
}
The most critical one lies in the accuracy of filter representation.  {
For harmonics based steerable filters \cite{weiler2018learning, weiler2019general},
 the harmonic bases are strongly bandlimited on purpose to be more robust to rotations (i.e., reducing aliasing during sampling, since the bases are defined in the continuous domain), which
usually leads to over-smooth configurations on filter representation as shown in Fig \ref{Fig:Filter}(b).
Moreover, proper radial profile is usually necessary for harmonic based filter parametrization,  e.g., improper parameter setting for the commonly used ring-shape radial profile would result to unexpected ring-shape artifacts. This also introduce a parameter tuning problem for radial profile.}
As for PDO-eConv, it is just designed for $5\times 5$ filters, and only contains 15 simple elementary filters \cite{shen2020pdo}.
Although it can well represent partial differential operators, it's not able to finely represent varying kinds and sizes of filters with high accuracy.
The aforementioned inaccuracy issue in current filter parametrization methods tends to affect the performance of the corresponding equivariant convolutions. Especially, this issue will be more severe in low-level image processing tasks, where the rotation equivariance of local features is important, but still has not been fully and accurately explored yet, calling for a practicable equivariant convolution method.

To address this issue,  {this study explores a filter parametrization method for equivariant convolution designing.} The contribution of this work can be mainly
summarized as follows:

1)  {We explore the filter parametrization method basing on 2D fourier series expansion, and propose a filter parametrization method which is able to alleviate the low-expression-accuracy problem and the aliasing effect simultaneously\footnote{ {The aliasing effect here is resulted from the insufficient sampling rate of the discrete
filter when the frequency of bases is too high, which will lead to an incorrect rotation result. Please refer to the supplementary material for more details and analysis.}}. Specifically, when the rotation degree is $\nicefrac{k\pi}{2}~(k\in\mathbb{N}$),
the proposed filter parametrization are exactly equivalent to 2D inverse discrete Fourier transform, which ensures that any discrete filter can be represented with no representation error. Besides, when the rotation degree is other than $\nicefrac{k\pi}{2}~(k\in\mathbb{N}$), as shown in Fig. \ref{Fig:Filter} (c) and (d), the proposed filter parametrization can largely alleviate the aliasing effect that usually occurs in conventional 2D Fourier series expansion, inclining to highly improve the representation capability of filter parametrization for both low-level and high-level tasks.}

2) With the proposed filter parametrization, we construct a new equivariant convolution method, named Fourier series expansion based equivariant convolution (F-Conv).
 {In F-Conv, we first time propose the rotation equivariant convolution for the output layer of CNNs, which is suitable for applying equivariant convolutions especially to low-level image processing tasks.}
We prove that the equivariance of F-Conv is exact in the continuous domain, while becoming approximate only after the discretization.
Moreover, we provide theoretical error analysis for the case when the equivariance is approximate, showing that the approximation error is dependent on the mesh size and filter size, complying with our common sense for this task.

3)  {Experiments on both high-level and low-level computer vision tasks have been implemented for evaluating the performance of the proposed method.
Particularly, this study applies filter parameterized equivariant convolutions to the low-level image super-resolution task for the first time along this research line, and achieves evidently better performance than conventional filter parametrization based equivariant convolutions.}
This reveals the possibility of filter parametrization based equivariant convolutions on the low-level image feature preservation, and the potential usefulness of this methodology to wider range of applications.

The paper is organized as follows. Section 2 reviews
the related works and introduces some necessary prior knowledge. Section 3 presents the proposed filter parametrization framework with theoretical analysis on its properties. Based on this filter parametrization method, Section 4 presents the F-Conv method on continuous functions and discrete domain, respectively. Section 5
then demonstrates experimental results implemented on typical high-level classification and low-level image super-resolution tasks, to substantiate the superiority of the proposed method both visually and quantitatively. The paper is finally
concluded with a future work discussion.

\section{Related Work and Prior Knowledge}

\subsection{Equivariant CNNs}

Early attempts for exploiting transformation symmetry prior in images are mainly designed by heuristics \cite{krizhevsky2012imagenet, laptev2016ti, esteves2017polar,sohn2012learning}. Data augmentation \cite{krizhevsky2012imagenet} is the most commonly used one among them.
The idea is to enrich the training set with transformed samples to train a model that is robust to the transformations.
Besides, TI-Pooling \cite{laptev2016ti} applied the transformation invariant pooling operator to the outputs of parallel architectures for the considered transformation set.
\cite{he2015delving,esteves2017polar} further transformed feature maps with differentiable modules to enforce equivariance transformations.
This category of approaches learns the transformation invariance directly from data, which, however, demands for a relatively large number of parameters and makes the network prone to overfitting.

 {
Recent works along this line focus on incorporating transformation equivariance directly into the network architecture, i.e., constructing equivariant convolutions. The aforementioned G-CNN \cite{Cohen2016group} and HexaConv \cite{hoogeboom2018hexaconv}  method  successfully construct equivariant convolutions for  $\nicefrac{\pi}{2}$ and $\nicefrac{\pi}{3}$ rotations, respectively. However, arbitrary degree  rotation equivariance convolutions remain a hard problem due to the discrete  filters exploited in these methods.}
%\cite{Cohen2016group} exploited the fact that $\nicefrac{\pi}{2}$ degree rotation and reflection on square lattice can be easily implemented
%by changing the position of elements in a filter, and proposed G-CNN with $\nicefrac{\pi}{2}$ degree rotation and reflection equivariant
%(i.e. $p4$ and $p4m$) on square lattices. By replacing the commonly used square lattices with hexagonal lattices,
%HexaConv succeeded in expanding rotation equivariant to $\nicefrac{\pi}{3}$ degree rotation ($p6m$) \cite{hoogeboom2018hexaconv}.
%Nevertheless, it is hard to make CNNs equivariant to the rotation angles other than $\nicefrac{\pi}{2}$ and $\nicefrac{\pi}{3}$ degrees in this intuitive way
%as \cite{Cohen2016group} and \cite{hoogeboom2018hexaconv} did, since it is not easy to design other rotational symmetric discrete lattices on the 2D plane.
In order to exploit more symmetries, \cite{Zhou2017Oriented} and \cite{Marcos2017Rotation} produced feature maps and filters at different orientations with bilinear interpolation, achieving an inherently approximately equivariant.
\cite{Worrall2017Harmonic} used harmonics to extract features and obtain 360$^\circ$ equivariance.
However, in these methods, the expected equivariance cannot be theoretically guaranteed after Gaussion-resampling or bilinear interpolation.

 {
Current equivariant CNN methods exploit filter parametrization technique for arbitrarily rotating filters in continuous domain.
\cite{weiler2018learning} and \cite{weiler2019general} make early attempts by employing harmonics as steerable filters to achieve exact equivariance with respect to
larger transformation groups in the continuous domain. The harmonic based approach guarantees full rotation equivariance which attracts practical application and theoretical research. Typically,  \cite{kondor2018generalization,cohen2019general} provided theoretical treatment of equivariant convolution, and derive generalized convolution formulaes.
Recently, \cite{shen2020pdo} and \cite{shen2021pdo} designed equivariance by relating convolutions with partial differential operators and proposed PDO-eConv, and firstly provided the error analysis to the approximation in the discrete domain.}

The key drawback of these filter parametrization methods is in the accuracy of filter representation. Specifically, for harmonics based method, the maximum angular frequencies for each radial part of harmonics are hard to choose, and one usually has to select lower angular frequencies to  {alleviate aliasing effect}, which will then easily lead to over-smooth issue.
Besides, the commonly used ring-shape radial parts need properly tuning multiple parameters,  it also can conduct unexpected ring-shape artifacts.
As for PDO-eConv, it is specifically designed for $5\times 5$ filters, with limited and relatively simple elementary filters \cite{shen2020pdo} which is hard to sufficiently represent varying kinds and sizes of filters with high accuracy.
Such inaccurate-representation issue might less affect the performance of high-level computer vision tasks, which mainly require relatively  coarse-scale transformation equivariance knowledge. However, for low-level problems, one has to consider to represent much finer-grained local details in pixel level with higher accuracy requirement, which makes current filter parametrization regimes hardly be used effectively.

\subsection{Prior Knowledge about Equivariance}

We follow the equivariance of prior works: Equivariance of a mapping transform means that a transformation on the input will result in a predictable transformation
on the output \cite{weiler2019general, shen2020pdo}.
Specifically, let $\Psi$ be a mapping from the input feature space to the output feature space, and $G$ is a group of transformations.
$\Psi$ is  equivariant with respect to the action of $G$, if for any $g\in G$,
\begin{equation}\label{equivariance}
   \Psi\left[ \pi_g[f] \right] = \pi_g'\left[\Psi[f]\right],
\end{equation}
where $f$ can be any input feature map in the input feature space, and $\pi_g$ and $\pi_g'$ denote how the transformation $g$ acts on input
and output features, respectively.

\begin{figure}[t]
\vspace{2mm}
\centering
\begin{center}
\includegraphics[width=1\linewidth]{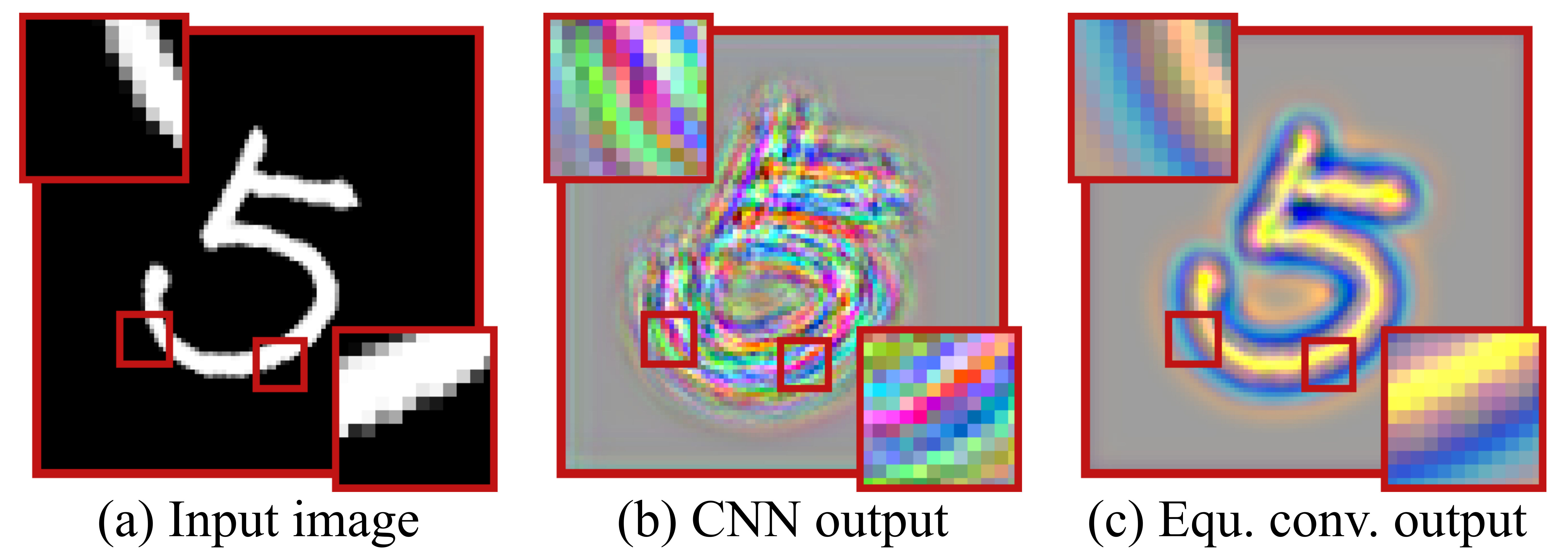}
\end{center}
\vspace{-5mm}
  \caption{{(a) A typical input cartoon image. (b)-(c) Outputs of randomly
initialized CNN and proposed rotation equivariant convolution network, respectively, where the demarcated areas
are zoomed in 3 times for easy observation.}}
\label{Fig:Eq}
\vspace{-1.5mm}
\end{figure}

In this paper, we focus on the rotation equivariance on 2D convolutions\footnote{The proposed method can be easily extended to rotation+reflection equivariance as previous works \cite{Cohen2016group, weiler2019general, shen2020pdo} did.}. In this case, $G$ is a group of rotation transformations,
and $\Psi$ is a convolution mapping.
{For local features, as shown in Fig. \ref{Fig:Eq}, the rotation equivariance can be easily understood as: Image local textures in different orientations will result to similar local features along corresponding orientations. From the figure, we can also observe that  rotation equivariant convolution is expected to better maintain the symmetry of local features underlying the image as compared to CNN.}

\section{Filter Parametrization Framework}\label{Filter Parametrization}

Filter parametrization is one of the most important concepts for the realization of equivariant convolution.
In this section, we explore the relationship between filter parametrization and Fourier series expansion, and conduct the proposed parametrization method.

\begin{figure*}[t]
\vspace{1mm}
\begin{center}
\hspace{-1mm}\includegraphics[width=1.0\linewidth]{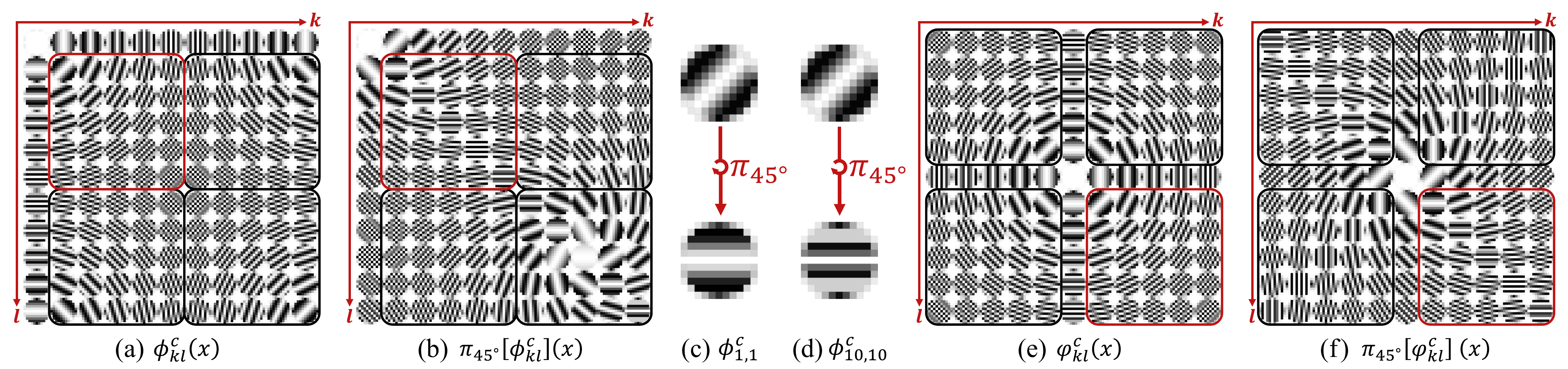}
\end{center}
\vspace{-5mm}
  \caption{{(a) Illustrations of the 2D Fourier bases $\phi^c_{kl}$, with $p=11, k,l = 0,1,\cdots,p-1$.
  The discretization of high frequency bases in black boxes are symmetrical to the low frequency bases in red box.
  (b) The $45^{\circ}$ rotation results of 2D Fourier bases.
  (c)-(d) illustration of rotating  $\phi^c_{1,1}$ and $\phi^c_{10,10}$ by $45^{\circ}$.
  Note that  $\phi^c_{1,1}$ is the same as $\phi^c_{10,10}$, but their rotation results are different to each other, which is due to the heavy aliasing effect in  $\phi^c_{10,10}$.
  (e)-(f) Illustrations of the proposed basis set (cosine part) and its $45^{\circ}$ rotation.  }}
\label{Fig:Method}
\vspace{-1.5mm}
\end{figure*}

%\subsection{Filter Parametrization Framework}
As shown in Fig. \ref{Fig:Filter}(a),
 the basic idea of filter parametrization is to define the objective functional filter as the linear combination of a set of basis functions
  $\{\psi_n\}_{n=1}^{N}$, aiming to get a learnable functional filter $\psi: \mathbb{R}^2\to\mathbb{R}$. Formally, it can be expressed as \cite{freeman1991design, weiler2018learning}:
\begin{equation}\label{Parametrization}
  \psi(x)=\sum_{n=1}^{N} w_n\psi_n(x),
\end{equation}
where $x=[x_1,x_2]^T\in\R^2$ denotes the 2 spatial coordinates, $N$ is the number of basis functions, $w_n$ is the $n$-th coefficient.
Moreover, the rotation operator $\pi_\theta$ can be easily expressed by coordinate transformation, that is:
\begin{equation}\label{Rotation}
  \pi_\theta[\psi](x) = \psi(U_\theta^{-1} x),
  \mbox{where}~ U_\theta \!=\!
  \begin{bmatrix}
  ~\cos\left(\theta\right) &  \sin\left(\theta\right)\\
  -\sin\left(\theta\right) & \cos\left(\theta\right)
  \end{bmatrix},
\end{equation}
where $\theta\in(-\pi, \pi]$ denotes the rotation angle and $\pi_\theta$ represents the $\theta$ degree rotation operator.
In practice, the rotation operator is usually adopted on the bases instead of the filter itself, that is
\begin{equation}\label{Rotated_Phi}
  \pi_\theta[\psi](x)=\sum_{n=1}^{N} w_n\pi_\theta[\psi_n](x)=\sum_{n=1}^{N} w_n\psi_n(U_\theta^{-1} x).
\end{equation}

Previous research has exploited different types of basis set for filter parametrization, such as harmonics,
partial-differential-operator-like and polynomial bases \cite{weiler2018learning,weiler2019general,shen2020pdo,xie2020color}.
To further improve the representation accuracy of filter parametrization, it is naturally to construct a Fourier series expansion based filter parametrization method.
Since  Fourier series expansion is equivariant to 2D inverse  discrete Fourier transform (DFT), which is with zero representation error.

\textbf{2D Fourier bases.}
 {A $p\times p$ discrete filter $\tilde{\phi}$ can be viewed as the discretization of an underlying 2D function $\phi(x)$,
sampling uniformly on the $\left[-\nicefrac{(p-1)h}{2},\nicefrac{(p-1)h}{2}\right]^2$ area of $\mathbb{R}^2$, where $h$ represents the mesh size of images.}
Formally, the 2D Fourier bases can be expressed as follows \cite{Brigham1988Fast}:
\begin{equation}\label{Fourierbase}
\begin{split}
   \phi_{kl}^c(x)&=\Omega(x)\cos\left(\frac{2\pi}{ph}[k,l]\cdot \begin{bmatrix}  x_1\\
  x_2  \end{bmatrix}\right), \\
  \phi_{kl}^s(x)&=\Omega(x)\sin\left(\frac{2\pi}{ph}[k,l]\cdot  \begin{bmatrix}  x_1\\
  x_2  \end{bmatrix}\right),
\end{split}
\end{equation}
where $k,l = 0,1,\cdots,p-1$ and $\Omega(x)\geq 0$ is a radial mask function\footnote{Please refer to the supplementary material for detailed settings of the radial mask function $\Omega(x)$.} that satisfies $\Omega(x) = 0$ if $\|x\|\geq(\nicefrac{p+1}{2})h$.
For intuitive understanding of Eq. (\ref{Fourierbase}), illustrations of Fourier bases (take cosine bases as example) are shown in Fig. \ref{Fig:Method}(a).
Then, the real part of 2D inverse DFT for $\tilde{\phi}$ (with circular shape mask) can be expressed in the following Fourier series expansion form:
\begin{equation}\label{Fourier}
\phi(x) = \sum_{k=0}^{p-1}\sum_{l=1}^{p-1}\left(a_{kl}\phi_{kl}^c(x) + b_{kl} \phi_{kl}^s(x) \right),
\end{equation}
where  $a_{kl}$ and $b_{kl}$ are expansion coefficients.  (\ref{Fourier}) is  a specific case of (\ref{Parametrization}),
which means it is also a filter parametrization.

Eq. (\ref{Fourier}) can represent any discrete filter $\tilde{\phi}$ (masked into circular shape) with zero representation
error\footnote{Zero representation error means $\tilde{\phi}_{kl}=\phi{(x_{kl})}$, where $x_{kl}$ is the coordinate to the element of $\tilde{\phi}$ in its $k$-th row and $l$-th column.}. This is because, one can set coefficients $a_{kl}$ and $b_{kl}$ as DFT results of $\tilde{\phi}$, and then inverse DFT can restore $\tilde{\phi}$ with zero error. Besides, it is easy to deduce that  $\Omega(x)$ doesn't affect the Fourier series expansion for $x$ s.t. $\Omega(x)\neq0$.
 Thus, Eq. (\ref{Fourier}) seems to be a rational filter parametrization.

{However, when we use this basis set to represent a rotated filter by Eq. (\ref{Rotated_Phi}), the rotation result is usually unsatisfying, as shown in Fig. \ref{Fig:Filter}(c).
The incorrect rotation result is due to aliasing effect\footnote{{More explanations on this issue is stated in supplementary file.}}, i.e., the sampling rate of the discrete filter is insufficient to the high  frequency bases.
Specifically, as shown in Fig. \ref{Fig:Method}(d), the aliasing effect will cause  the Fourier bases with higher frequencies to be badly destroyed after rotation by (\ref{Rotation}), which effect less to the bases with low frequencies (as shown in Fig. \ref{Fig:Method}(c)).
These aliasing effects tend to seriously hamper the accuracy for representing the rotated filter, and thus such 2D Fourier bases set should not be properly suggested to be directly employed for designing filter parametrization methods.}

\textbf{Proposed bases and filter parametrization.} {The key issue now is to alleviate the aliasing effect in Fourier series expansion for the rotated cases, while possibly keep its high accuracy for the rotation free cases.
%Firstly, let us have a look at the 1D example depicted in Fig. \ref{Fig:Method} (e) for easily understanding the aliasing effect.  {From the figure, we can see that the discretizations of a high frequency and a low frequency cosine function can be exactly the same. However, the discretization  of the transformation results of the high frequency one tend to be highly unpredictable.
%Such aliasing effect also easily occurs in the commonly used 2D Fourier series expansion.
As clearly depicted by the bases in the lower right corner of Fig. \ref{Fig:Method} (a) and (b), 2D Fourier bases with high frequencies will be badly destructed after rotation.
Fortunately, these high frequency bases are actually symmetry to those with low frequencies (in the top left of Fig. \ref{Fig:Method}(a)) when there are no rotations. This  symmetry
between high frequency and low frequency bases can be used for alleviating aliasing effect.}

Specifically, as shown in Fig. \ref{Fig:Method}, the bases of the four areas (marked with boxes in red and black colors) in Fig. \ref{Fig:Method} (a) are symmetrical.
This means that we can use the mirror functions of low frequency bases (in the red boxes) to replace the high frequency bases (in the black boxes, which intrinsically cause the aliasing effect).  { In other words, for all $k>p-1-\left\lfloor\nicefrac{p}{2}\right\rfloor$, we should use
a basis with frequency $k-p$ to replace the basis with frequency $k$ ($\left\lfloor\cdot\right\rfloor$ is the floor  operator). This will change the value of frequency from $\{0,1,\cdots, p-1\}$ to $\{-\left\lfloor\nicefrac{p}{2} \right\rfloor\,\cdots, 0,\cdots,p-1-\left\lfloor\nicefrac{p}{2} \right\rfloor\}$.
Formally, we  propose  the following bases:
\begin{equation}\label{Proposed_Bases}
\begin{split}
  \varphi_{kl}^c(x)&=\Omega(x)\cos\left(\frac{2\pi}{ph}\left[k-\left\lfloor\frac{p}{2} \right\rfloor,l-\left\lfloor\frac{p}{2} \right\rfloor\right]\cdot \begin{bmatrix}  x_1\\
  x_2  \end{bmatrix}\right),\\
  \varphi_{kl}^s(x)&=\Omega(x)\sin\left(\frac{2\pi}{ph}\left[k-\left\lfloor\frac{p}{2} \right\rfloor,l-\left\lfloor\frac{p}{2} \right\rfloor\right]\cdot \begin{bmatrix}  x_1\\
  x_2  \end{bmatrix}\right),
\end{split}
\end{equation}
where $k,l = 0,1,\cdots, p-1$, and $\Omega(x)$ is the aforementioned radial mask function. } Illustrations of these bases are shown in Fig. \ref{Fig:Method}(a) for easy visualization\footnote{ {Since $\forall k,l\in\mathbb{Z}$, $\cos\left(v(kx_1+lx_2)\right)=\cos\left(v(-kx_1-lx_2)\right)$  and $\sin\left(v(kx_1+lx_2)\right)=-\sin\left(v(-kx_1-lx_2)\right)$,  we actually only need about half number of the  bases in Eq. (\ref{Proposed_Bases}). Especially, when $p$ is a odd number, $p^2$ bases are enough. More detailed analysis on the proposed bases have been made in the supplementary material.}}.

To analyze the properties of the proposed bases,
we first deduce the following conclusion\footnote{The proofs of all the theoretical results in this paper are presented in supplementary material due to page limitation. }:

\begin{Rem}
For any mesh size $h\in\mathbb{R}$,  filter size $p\in \mathbb{N}_{+}$, and  grid point  $x$  on the $p\times p$ mesh of~$[\nicefrac{(1-p)h}{2}, \nicefrac{(p-1)h}{2}]^2$,
i.e., $x_1 = \left(i-\nicefrac{(p-1)}{2}\right)h$, $x_2 = \left(j-\nicefrac{(p-1)}{2}\right)h$, $\forall i, j  = 0,1,\cdots, p-1$, let  $k, l  = 0,1,\cdots, p-1$, and then it holds that,
\begin{equation}\label{Theorem1}
\begin{split}
  \phi^c_{kl}(x) &= s(k,l)\cdot \varphi^c_{\mathcal{I}(k),\mathcal{I}(l)}(x), \\
   \phi^s_{kl}(x) &= s(k,l)\cdot \varphi^s_{\mathcal{I}(k),\mathcal{I}(l)}(x),
\end{split}
\end{equation}
where $\varphi^c_{kl}$ and $\varphi^s_{kl}$ are defined in (\ref{Proposed_Bases}),
$\phi^c_{kl}$, $\phi^s_{kl}$ are defined in (\ref{Fourierbase}), $\mathcal{I}(\cdot) = \left((\cdot)+\left\lfloor\nicefrac{(p-1)}{2} \right\rfloor\right)\%p$,
%$\forall m = 0,1,\cdots,p-1, \mathcal{I}(m)\in\{0,1,\cdots,p-1 \}$, satisfying
%\begin{equation}\label{IP}\nonumber
%\mathcal{I}(m) =\left\{
%  \begin{matrix}
%  \left\lfloor\nicefrac{(p-1)}{2} \right\rfloor + m & \mbox{if} &  m\leq \nicefrac{p}{2}\\
%  \left\lceil\nicefrac{(p-1)}{2} \right\rceil - m  & \mbox{if} &  m > \nicefrac{p}{2}
%  \end{matrix}\right.,
%\end{equation}
and $s(k,l)\in \{-1,1\}$, satisfying $s(k,l) = \emph{\mbox{sign}}(k-\nicefrac{p}{2}+\epsilon)^{p-1}\cdot\emph{\mbox{sign}}(l-\nicefrac{p}{2}+\epsilon)^{p-1}$, $0<\epsilon<\nicefrac{1}{2}$.
\end{Rem}

From {Remark 1}, it is easy to see that, the proposed basis set is exactly equivalent to commonly used 2D Fourier bases when there is no rotation (or other transformations),
just some of their expansion coefficients being with opposite signs when $s(k,l)\!=\!-1$. Especially, when $p$ is an odd number,
as shown in Fig. \ref{Fig:Method} (a) and (e), it is easy to observe that the framed parts in these two figures are the same to each other. This implies that the filter parametrization so designed is equivalent to inverse 2D DFT when there is no rotation, and there is no representation error in this case.

Moreover, as shown in Fig. \ref{Fig:Method}(f), the aliasing effect can be largely alleviated when we rotate the proposed bases.
Note that the smallest period of the commonly used 2D Fourier bases is $\frac{p}{p-1}h\approx h$.
Thus, that there is a non-zero possibility for one peak and one valley to exist simultaneously between two grid points,
which intrinsically leads to heavy aliasing effect in rotation results as shown in Fig. \ref{Fig:Method} (b) and (c).
Comparatively, the smallest period of the proposed bases is $\left(\nicefrac{ph}{\left\lfloor f_0 \right\rfloor}\right)\geq {2h}$.
There is thus at most one peak or valley between two grid points\footnote{Half period of cosine and sine functions can only contain one peak or valley. },
 which helps alleviate the aliasing effect.

According to the above analysis, the proposed filter parametrization not only faithfully keeps the high accuracy of 2D Fourier bases, but also well weakens the issue of the aliasing effect.
In the following, we will introduce the equivariant convolutions based on this basis set.

\section{Equivariant Convolution Framework}
%The input and intermediated layers of equivariant convolutions have been while defined in previous works \cite{weiler2018learning, weiler2019general,shen2020pdo}, and theoretically proved to be exactly equivariant in continuous domain. In this section, we further introduce a novel equivariant convolution for output layers, which is necessary for low-level computer vision tasks.

\begin{table}
  \caption{The involved concepts and notations for equivariant convolutions (eConv) in the continuous and discrete domains, respectively.  } \vspace{-2mm}
  \label{Notation-table}
  \centering \setlength{\tabcolsep}{13pt}
  \begin{tabular}{lcc}
    \toprule
    \multirow{2}{*}{Concept}     &\multicolumn{2}{c}{Notation}   \\
    \cmidrule(r){2-3}
             &   Continuous    & Discrete \\
    \midrule
    Input Image & $r(x)$       & $I$             \\
    Transformation Group &      $O(2)$         &  $S$           \\
    Group element/Index&      $A,B \in O(2)$  &  $A,B \in S$                            \\
    Feature Map & $e(x,A)$  & $F^A$  \\
    \midrule
    Filter (Input) & $ \varphi_{in}\left(A^{-1}x\right) $     &   $\tilde{\Psi}^A$ \\
    Filter (Intermediate) & $ \varphi_{A}\left(B^{-1}x\right) $     &   $\tilde{\Phi}^{B,A}$ \\
    Filter (Output) & $ \varphi_{out}\left(B^{-1}x\right) $     &   $\tilde{\Upsilon}^B$ \\
    \midrule
    eConv (Input) & $\Psi[r]$ &  $\tilde{\Psi} \star I$ \\
    eConv (Intermediate) & $\Phi[e]$ &  $\tilde{\Phi} \star F$ \\
    eConv (Output) & $\Upsilon[e]$ &  $\tilde{\Upsilon} \star F$ \\
    \bottomrule
  \end{tabular}\vspace{-1.5mm}
\end{table}

 {In this section, we first introduce the equivariant convolutions in continuous domain. It should be noted that the input and intermediated layers of equivariant convolutions have been well defined in previous works \cite{weiler2018learning, weiler2019general,shen2020pdo}, while we further introduce a novel equivariant convolution for the output layer, which is necessary for low-level computer vision tasks.
Then, we discretize the proposed equivariant convolutions, and provide theoretical analysis for the equivariant error in discrete domain.}

\subsection{ {Necessary Notations and Concepts}}
To avoid possible confusion cased by different notations in continuous and discrete domains, we list the major notations correspondingly used in two domains in Table \ref{Notation-table}.

Specifically, we represent an image as a two-dimensional grid function $I \in\R^{n\times n}$, and represent an intermediate feature map as $F$. Note that in rotation equivariant networks, a feature map is a multi-channel matrix (i.e., a tensor),   as shown in Fig. \ref{Fig:eConv} (a), with $F\in\R^{n\times n\times t}$, where the third mode is  with respect to the rotation group $S$, and $t$ is the number of elements in $S$. Moreover,
we denote a specific channel in $F$ as $F^A\in\R^{n\times n}$,  where $A\in S$ is a rotation matrix, and also  used as an index for denoting a specific channel in $F$.

 {In the continuous domain, we follow the previous works \cite{weiler2019general,shen2020pdo}, and consider the equivariance on the orthogonal group  $O(2)$\footnote{The rotation group  $S$ represents a subgroup of $O(2)$, and it is also regarded as the discretization of $O(2)$ in this paper.}. Formally,
$O(2)=\{A\in\mathbb{R}^{2\times 2}|A^TA = I_{2\times  2}\}$, which contains all rotation and reflection matrices. Without ambiguity, we use $A$ to parameterize $O(2)$. We consider the Euclidean group $E(2) = \mathbb{R}^2 \rtimes O(2)$ ($\rtimes$ is a semidirect-product), whose element is represented as $(x, A)$. Restricting the domain of $A$ and $x$, we can also use this representation to parameterize any subgroup of $E(2)$.
The  input image can be modeled as a function defined on $\mathbb{R}^2$, denoted as $r(x)$. The intermediate feature map can be model as a function defined on $E(2)$, denoted as $e(x,A)$ (as shown in Table \ref{Notation-table}, and $r(x)$ and $e(x,A)$ are the continuous versions of $I$ and $F^A$, respectively).
We denote the function spaces of $r$ and $e$ as $C^\infty(\mathbb{R}^2)$ and $C^\infty(E(2))$,
respectively\footnote{The smoothness of e means that the feature map $e(x,A)$ is smooth with respect to $x$ when $A$ is fixed.
For simplicity, we set the functional space as $C^\infty(\mathbb{R}^2)$. Actually,  in implementation, we only require that $r\in C^2(\mathbb{R}^2)$.
The requirement on $e$ is the same.}.}

 {With above notations,  the transformations on the input and feature maps can be mathematically formulated. Specifically,
for an input $r\in C^\infty(\mathbb{R}^2)$ and transformation $\tilde{A}\in O(2)$,  $\tilde{A}$ acts on $r$ by
\begin{equation}\label{piR}
  \pi^R_{\tilde{A}}[r](x) = r(\A^{-1}x), \forall x\in\mathbb{R}^2.
\end{equation}
For a feature map $e\in C^\infty(E(2))$ and transformation $\tilde{A}\in O(2)$, $\tilde{A}$ acts on $e$ by
\begin{equation}\label{piE}
  \pi^E_{\tilde{A}}[e](x,A) = e(\A^{-1}x,\A^{-1}A), \forall (x,A)\in E(2).
\end{equation}
These two transformations will be used in the theoretical analysis.
}

\subsection{Equivariant Convolutions in Continuous Domain}\label{ContinuousSection}

\begin{figure*}[t]
\vspace{-1mm}
\begin{center}
\hspace{-1mm}\includegraphics[width=1.0\linewidth]{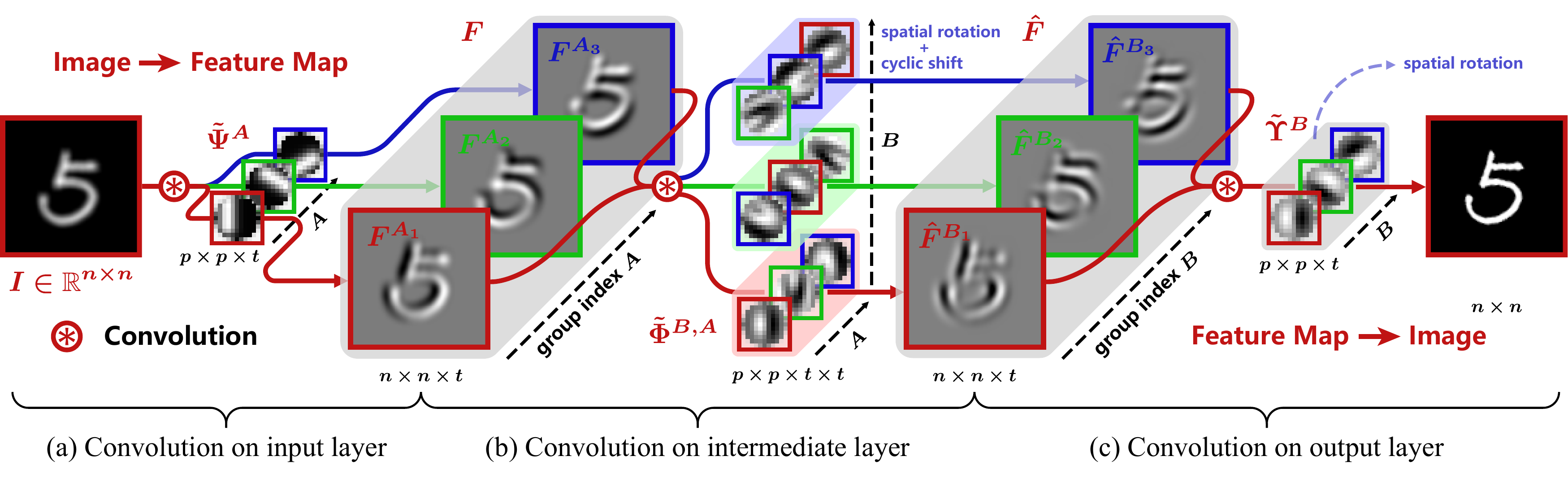}
\end{center}
\vspace{-6.5mm}
  \caption{Illustration of an example network constructed by the proposed equivariant convolutions, where we set the  transformation group $S$ ($A_i, B_i\in S$) as $\nicefrac{2\pi i}{3}$ rotations, $i = 1,2,3$. (a)-(c) Equivariant convolutions  of the input layer, intermediate layers, and output layer, respectively.  }
\label{Fig:eConv}
\vspace{-1.5mm}
\end{figure*}

 {The equivariant convolutions for input and intermediate layers have been well defined in previous works \cite{weiler2019general, weiler2018learning, kondor2018generalization,cohen2019general}, which can be expressed in following formulations.}

\textbf{Input layer.} The convolution for input layer is denoted as $\Psi$,
which maps an input $r\in C^\infty(\R^2)$ to a feature map defined on $E(2)$. Specifically, for any $(y, A)\in E(2)$,
\begin{equation}\label{input_Conv}
  \Psi[r](y,A) =  \int_{\R^2}\varphi_{in}\left(A^{-1}x\right)r(y-x)d\sigma(x),
\end{equation}
where $\sigma$ is a measure on $\R^2$ and $\varphi_{in}$ is a parameterized filter defined in the formulation of Eq. (\ref{Parametrization}).
%Note that (\ref{input_Conv}) can be roughly viewed as convoluting $r(x)$ with a set of transformed filters, $\left\{\varphi_{in}\left(A^{-1}x\right)| A\in O(2)\right\}$.

\textbf{Intermediate layers.} The convolution for intermediate layer is denoted as $\Phi$,
which maps an feature map $e\in C^\infty(E(2))$ to another feature map defined on $E(2)$. Specifically, for any $(y, B)\in E(2)$,
\begin{equation}\label{regular_Conv}
\begin{split}
   \Phi[e](y,\!B) \!\!=\!\!\!
   \int_{\!O(2)}\!\int_{\R^2}\!\!\varphi_{\!A}\!\left(B^{-\!1}x\right)\!e(y\!-\!x, \!B\!A)d\sigma(x)dv(A),
\end{split}
\end{equation}
 {where $v$ is the Haar measure on $O(2)$}, $A,B\in O(2)$ denote orthogonal transformations in the considered group,
and $\varphi_{A}$ indicates the filter with respect to the channel of feature map indexed by $A$. Note that the discrete version of these convolutions are shown in Fig. \ref{Fig:eConv} for easy understanding.

 {In practice, different manners have been proposed to extract the information of interest in the final output layer of networks.
For high-level tasks, such as  classification and segmentation,
the output layer for rotation-invariant networks can be easily set as a pooling over the orientation dimension \cite{weiler2018learning,weiler2019general,shen2020pdo}.
Therefore, equivariant convolutions for input and intermediate layers are enough for these tasks.
However, for low-level tasks, an equivariant convolution for the output layer is supplementally required to guarantee
equivariance while avoiding the detail texture losing of images. We thus define the following equivariant convolution for the output layer.}

\textbf{Output layer.} We use $\Upsilon$ to denote the convolution on the final layer,
which maps a feature map $e\in C^\infty(E(2))$ to a function in $C^\infty(\R^2)$. Specifically, for any $y\in \R^2$, we define:
\begin{equation}\label{output_Conv}
  \Upsilon[e](y) \!=\! \!\int_{O(2)} \int_{\R^2}\!\!\varphi_{out}\!\left(B^{-1}x\right)e(y\!-\!x, B)d\sigma(x)dv(B),
\end{equation}
where $B\in O(2)$ and $\varphi_{out}$ is a parameterized filter.

 {The convolutions for input and intermediate layers have been proved in previous work \cite{weiler2018learning} that they  are equivariant under orthogonal transformations ($O(2)$). By using similar manner, it is easy to deduce that the proposed  convolution for output layer is also equivariant under orthogonal transformations.
In summery, we have the following  conclusion.}

\begin{Rem} For $r\in C^\infty(\R^2)$, $e\in C^\infty(E(2))$ and $\tilde{A}\in O(2)$,
the following results are satisfied:
\begin{equation}\label{Thm2}
\begin{split}
  \Psi\left[\pi_{\tilde{A}}^R\left[r\right]\right] &= \pi^E_{\tilde{A}}\left[\Psi\left[r\right]\right],\\
  \Phi\left[\pi_{\tilde{A}}^E\left[e\right]\right] &= \pi^E_{\tilde{A}}\left[\Phi\left[e\right]\right],\\
  \Upsilon\left[\pi_{\tilde{A}}^E\left[e\right]\right] &= \pi^R_{\tilde{A}}\left[\Upsilon\left[e\right]\right],\\
\end{split}
\end{equation}
where $\pi_{\tilde{A}}^R$, $\pi_{\tilde{A}}^E$, $\Psi$, $\Phi$ and $\Upsilon$
are defined by (\ref{piR}), (\ref{piE}), (\ref{input_Conv}), (\ref{regular_Conv}) and (\ref{output_Conv}), respectively.
\end{Rem}

By substituting the filter parametrization as defined in Eq. (\ref{Parametrization}), we can obtain the proposed equivariant convolutions, i.e., Fourier series expansion based equivariant convolution (F-Conv).
Since the convolution operators are naturally translation equivariant, it is easy to verify that the proposed convolutions are
equivariant over $E(2)$.
%Combining with nonlinearities such as ReLU, which do not disturb the equivariance, we can construct networks that preserves equivariance over the entire network.

\subsection{Equivariant Convolutions on Discrete Domain}\label{DiscreteSecion}

Next, we show how to apply the F-Conv to 2D digital images.
Formally,  an image $I \in R^{n\times n}$ can be viewed as a two-dimensional
 discretization of a smooth function $r: \R^2\to\R$ at the cell-center of a regular grid with $n\times n$ cells, i.e., for $i, j = 1,2, \cdots, n$,
\begin{equation}\label{I}
  I_{ij} = r(x_{ij}),
\end{equation}
where $x_{ij} \!=\! \left(\left(i\!-\!\frac{n\!+\!1}{2}\right)h, \left(j\!-\!\frac{n\!+\!1}{2}\right)h\right)^T\!$ and  $h$ is the mesh size.

Similarly, an intermediate feature map $F\in \R^{n\times n \times t}$ in equivariant networks is a multi-channel  tensor, which can be viewed as the discretization of a continuous function defined on $\tilde{E} = \R^2\rtimes S$, where $S$ is a subgroup\footnote{In practice, the subgroup is usually assumed to contain $t$ rotations with $\nicefrac{2\pi}{t}$ degree for an integer $t\in \mathbb{N}_+$. } of $O(2)$ and $t$ is the number of elements in $S$. Formally,   $F$ can be represented as a three-dimensional grid tensor sampled from a smooth function $e: \R^2\times S\to \R$, i.e., for $i, j = 1,2, \cdots, n$,
\begin{equation}\label{F}
  F_{ij}^A = e(x_{ij}, A),
\end{equation}
where $x_{ij} = \left(\left(i-\frac{n+1}{2}\right)h, \left(j-\frac{n+1}{2}\right)h\right)^T$ and $A\in S$.

A single channel $p\times p$ filter represents a two-dimensional grid function obtained by discretizing a smooth function $\varphi: \R^2 \to \R$, which satisfies $\varphi(x) = 0, \forall x, s.t., \|x\|\geq(\nicefrac{p+1}{2})h$. Accordingly, we define filters for input, intermediate and output layers as $\tilde{\Psi}\in \R^{p\times p \times t}$, $\tilde{\Phi}\in \R^{p\times p \times t \times t}$ and $\tilde{\Upsilon}\in \R^{p\times p \times t}$, respectively, where $t$ is the number of elements in $S$. For $i, j = 1,2, \cdots, p$, and $A,B\in S$, we denote
\begin{equation}\label{Phi_d}
\begin{split}
  &\tilde{\Psi}_{ij}^A = \varphi_{in}\left(A^{-1}x_{ij}\right),\\
 &\tilde{\Phi}_{ij}^{B,A} = \varphi_A\left(B^{-1}x_{ij}\right),\\
  &\tilde{\Upsilon}_{ij}^{B} = \varphi_{out}\left(B^{-1}x_{ij}\right),
\end{split}
\end{equation}
where $x_{ij} = \left(\left(i-\frac{p+1}{2}\right)h, \left(j-\frac{p+1}{2}\right)h\right)^T$, $\varphi_{in}$, $\varphi_{out}$ and $\varphi_A$ are parameterized filters   defined in the formulation of (\ref{Parametrization}), and $\varphi_{{A}}$ indicates the filter with respect to the channel of feature map indexed by ${A}$.

It should be noted that  the filters in (\ref{Phi_d}) can be learnt by calculating the  expansion coefficients as shown in (\ref{Parametrization}). For example, for any $ A\in S$, $\tilde{\Psi}^A$ shares a set of representation coefficients, and
can be obtained by learning the coefficients of the bases $\left\{\tilde{\varphi}^{cA}_{kl}, \tilde{\varphi}^{sA}_{kl}|k,l = 0,1,\cdots,p-1\right\}$, where
\begin{equation}\label{bases_d}
\begin{split}
   \left(\tilde{\varphi}^{cA}_{kl}\right)_{ij}=\varphi^c_{kl}(A^{-1}x_{ij}),\left(\tilde{\varphi}^{sA}_{kl}\right)_{ij}=\varphi^s_{kl}(A^{-1}x_{ij}).
\end{split}
\end{equation}

Accordingly, we can discretize the  continuous convolutions for input layer, intermediate layers and output layer (i.e., (\ref{input_Conv}), (\ref{regular_Conv}) and (\ref{output_Conv}),) as follows.

\textbf{Input layer.} For any $A\in S$  the convolution  of $\tilde{\Psi}$ and $I$ can be computed by discretizing Eq. (\ref{input_Conv}), which is
\begin{equation}\label{input_2}
\left(\tilde{\Psi}\star I\right)^A =\tilde{\Psi}^A*I,
\end{equation}
where $*$ is the commonly used discrete 2D convolution. %One can see Fig \ref{Fig:eConv} (a) for easy understanding this convolution.

\textbf{Intermediate layers.} For any $ B\in S$ the convolution of $\tilde{\Phi}$ and $F$ can be computed by discretizing Eq. (\ref{regular_Conv}), which is
\begin{equation}\label{regular_2}
  \left(\tilde{\Phi}\star F\right)^B =\sum_{A\in S}\tilde{\Phi}^{B,A}*F^{BA}.
\end{equation}
The implementation of (\ref{regular_2}) is relatively more complex, since there is a channel shift of $F$ caused by adopting $B$ on its channel indexes.  Therefore,  we will further introduce the implementation detail of this convolution in the later sections. So far, one can see Fig \ref{Fig:eConv} (b) for a sketchy understanding of this equivariant convolution.

\textbf{Output layer.}  The convolution of $\tilde{\Upsilon}$ and $F$ can be computed by discretizing Eq. (\ref{output_Conv}), which  is
\begin{equation}\label{output_2}
  \tilde{\Upsilon}\star F =\sum_{B\in S}\tilde{\Upsilon}^{B}\ast F^{B}.
\end{equation}
One can see Fig \ref{Fig:eConv} for easy understanding these equivariant convolutions.

Although it has been  proved that the equivariance of continuous convolutions (\ref{input_Conv}), (\ref{regular_Conv}) and (\ref{output_Conv}) are exact,
their discretization still contain approximation errors. We then provide an error analysis for the discretized convolutions.
Firstly, we denote the transformations on $I$ and $F$ by
\begin{equation}\label{Pi_d}
\begin{split}
    & \left(\tilde{\pi}_{\A}^{R}(I)\right)_{ij} = \pi_{\A}^{R}[r](x_{ij}),  \left(\tilde{\pi}_{\A}^{\E}(F)\right)_{ij}^A = \pi_{\A}^{E}[e](x_{ij},A),\\
    &\forall i,j = 1,2,\cdots,n, \forall A,\A \in S.
\end{split}
\end{equation}
Then, we deduce the following theorem for evaluating approximation errors of the equivariance.

\begin{Thm} Assume that an image $I\in \R^{n\times n}$ is discretized from the smooth function $r:\R^2\to\R$ by (\ref{I}), a feature map $F\in \R^{n\times n \times t}$ is discretized from the smooth function $e:\R^2\times S\to\R$ by (\ref{F}), $|S|=t$, and filters $\tilde{\Psi}$, $\tilde{\Phi}$ and $\tilde{\Upsilon}$ are generated from $\varphi_{in}$, $\varphi_{out}$ and $\varphi_{A}, \forall A\in S$, by (\ref{Phi_d}), respectively. If for  any $A\in S,~ x\in \R^2$, the following conditions are satisfied:
\begin{equation}\label{condition}
  \begin{split}
      &|r(x)| , |e(x,A)|\leq F_1,\\
            &\|\nabla r(x)\| , \|\nabla e(x,A) \|\leq G_1,\\
                  &\|\nabla^2 r(x) \| , \|\nabla^2 e(x,A) \|\leq H_1,\\
      &|\varphi_{in}(x)|, |\varphi_{A}(x)|, |\varphi_{out}(x)|\leq F_2, \\
       &\|\nabla \varphi_{in}(x) \|, \|\nabla \varphi_{A}(x) \|, \|\nabla \varphi_{out}(x) \|\leq G_2, \\
       &\|\nabla^2 \varphi_{in}(x)\|, \|\nabla^2 \varphi_{A}(x) \| , \|\nabla^2 \varphi_{out}(x) \|\leq H_2, \\
      &\forall \|x\|\geq\nicefrac{(p+1)h}{2},~ \varphi_{in}(x), \varphi_{A}(x), \varphi_{out}(x)  =0,
  \end{split}
\end{equation}
where $p$ is the filter size, $h$ is the mesh size, and $\nabla$ and ${\nabla}^2$ denote the operators of gradient and Hessian matrix, respectively, then for any $\A\in S$, the following results are satisfied:
\begin{equation}\label{Thm3}
\begin{split}
   &\left\|\tilde{\Psi}\star\tilde{\pi}_{\tilde{A}}^R\left(I\right) - \tilde{\pi}^{\E}_{\tilde{A}}\left(\tilde{\Psi}\star I \right)\right\|_{\infty}
    \leq\frac{C}{2}(p+1)^2h^2  , \\
   &\left\|\tilde{\Phi}\star\tilde{\pi}_{\tilde{A}}^{\E}\left(F\right) - \pi^{\E}_{\tilde{A}}\left(\tilde{\Phi}\star F\right)\right\|_{\infty}
    \leq\frac{C}{2}(p+1)^2h^2t  ,\\
   &\left\|\tilde{\Upsilon}\star \tilde{\pi}_{\tilde{A}}^{\E}\left(F\right) - \tilde{\pi}^R_{\tilde{A}}\left(\tilde{\Upsilon}\star F\right)\right\|_{\infty}
    \leq\frac{C}{2}(p+1)^2h^2t  ,
\end{split}
\end{equation}
where $C = F_1H_2+F_2H_1+2G_1G_2$, $\tilde{\pi}_{\tilde{A}}^R$, $\tilde{\pi}_{\tilde{A}}^{\E}$, $\tilde{\Psi}$, $\tilde{\Phi}$ and $\tilde{\Upsilon}$
are defined by (\ref{Phi_d}) and (\ref{Pi_d}), respectively, the operators $\star$ involved in Eq. (\ref{Thm3}) are defined in (\ref{input_2}), (\ref{regular_2}) and (\ref{output_2}), respectively, and $\|\cdot\|_{\infty}$ represents the infinity norm.
\end{Thm}

In practice, the conditions in the above theorem are easy to be satisfied, which only need the first and second derivatives of the underlying input function to be bounded. From the theorem, it is easy to see that the accuracy of equivariance is mainly dependent on the patch size, the mesh size and the group size. This fully complies with our common sense that the smaller mesh size and the fewer pixel number in a filter are, the smaller approximation error should be. When the mesh size approaches zero, the approximation error also approaches zero.
Moreover, the theorem also implies that applying a rotation to the input image results in a joint spatial rotation operation and cyclic shift over the orientation indices of the feature maps.

Combining the proposed discrete F-Conv with nonlinearities such as ReLU, which do not disturb the equivariance, we can construct networks that preserve equivariance over the entire network.

\subsection{Implementation Details }

\textbf{Implementation of group convolutions.}
The proposed convolutions for the input and the output layers are easy to be implemented in practice, since (\ref{input_2})  and  (\ref{output_2}) are both with the expressions of commonly used convolution in popular softwares. However, the convolution for the intermediate layers  (\ref{regular_2}) is relatively complex, for there is a $B$ adopted on the channel indexes of $F$. Fortunately, previous works  \cite{weiler2018learning,weiler2019general,shen2020pdo} have introduced an efficient manner for fastly implementing it.
Specifically, we first rewrite (\ref{regular_2}) as
 \begin{equation}\label{regular_3}
  \left(\tilde{\Phi}\star F\right)^B  = \sum_{A\in S}\tilde{\Phi}^{B,B^{-1}A}*F^{A}.
\end{equation}
Then, it is easy to see that  Eq. (\ref{regular_3}) can be implemented by performing the commonly used convolution on $F$ and  a ${p\times p\times t\times t}$ filter rotated spatially and shifted cyclically along the third mode, as shown in Fig. \ref{Fig:eConv}(b).  More specifically speaking, we prepare a filter  $\bar{\Phi}\in\R^{p\times p\times t\times t}$, which satisfies $\bar{\Phi}^{B,A} = \tilde{\Phi}^{B,B^{-1}A} $. Then, Eq. (\ref{regular_3}) can be performed by convoluting  $\bar{\Phi}$ and $F$.

\textbf{Normalization of bases in F-Conv.}  Formally, a  set of rotated filters can be represented as  $\tilde{\Psi}\in\R^{p\times p \times t}$, where $t$ is the number of orientations. Let $\varphi_n$ be the $n$-th element of $\left\{\varphi^c_{kl},\varphi^s_{kl}|k,l = 1,2,\cdots,p \right\}$ and  $w_n$ be the $n$-th element of $\left\{a_{kl},b_{kl}|k,l = 1,2,\cdots,p \right\}$, and then the parametrization of $\tilde{\Psi}$ can be represented as $\tilde{\Psi}^A_{ij} = \sum_n w_n\varphi^A_n =\sum_n w_n\varphi_n(A^{-1}, x_{ij})$. Then, we can obtain
\begin{equation}\label{ParameterRotated}
  \mbox{vec}({\tilde{\Psi}}) = \left[\begin{matrix}\mbox{vec}(\varphi_1^{A_1})\\ \vdots\\\mbox{vec}(\varphi_1^{A_t}) \end{matrix} \begin{matrix}\cdots\\ \ddots \\\cdots \end{matrix}\begin{matrix}\mbox{vec}(\varphi_n^{A_1})\\ \vdots\\\mbox{vec}(\varphi_n^{A_t}) \end{matrix} \right] \!\cdot\! \begin{bmatrix}w_1\\ \vdots\\w_n \end{bmatrix}\triangleq Dw,
\end{equation}
where $\mbox{vec}(\cdot)$ is the vectorization operator, $D\in \R^{tp^2\times 2p^2}$.  In practice, we perform singular value decomposition on $D$, i.e.,
$D = U \Sigma V^T$, and replace $D$ with $U\in \R^{tp^2\times r}$, where $r$ is the rank of $D$, and represent the parameterized filter by:
\begin{equation}\label{Ur}
  \mbox{vec}({\tilde{\Psi}}) = U\hat{w},
\end{equation}
where $\hat{w} = \Sigma V^Tw$ is the new coefficients. This can reduce the redundant parameters and  {make the learning of parameters easier}. It can also avoid the non-uniqueness of to-be-estimated coefficients, which is caused by the redundancy of columns in $D$. Besides, by Eq. (\ref{Ur}), $\|\tilde{\Psi}\| = \|\hat{w}\|$, and this will lead to an easy initialization  scheme on $\hat{w}$.

\begin{figure*}[t]
\vspace{0mm}
\hspace{-0mm}\includegraphics[width=1.0\linewidth]{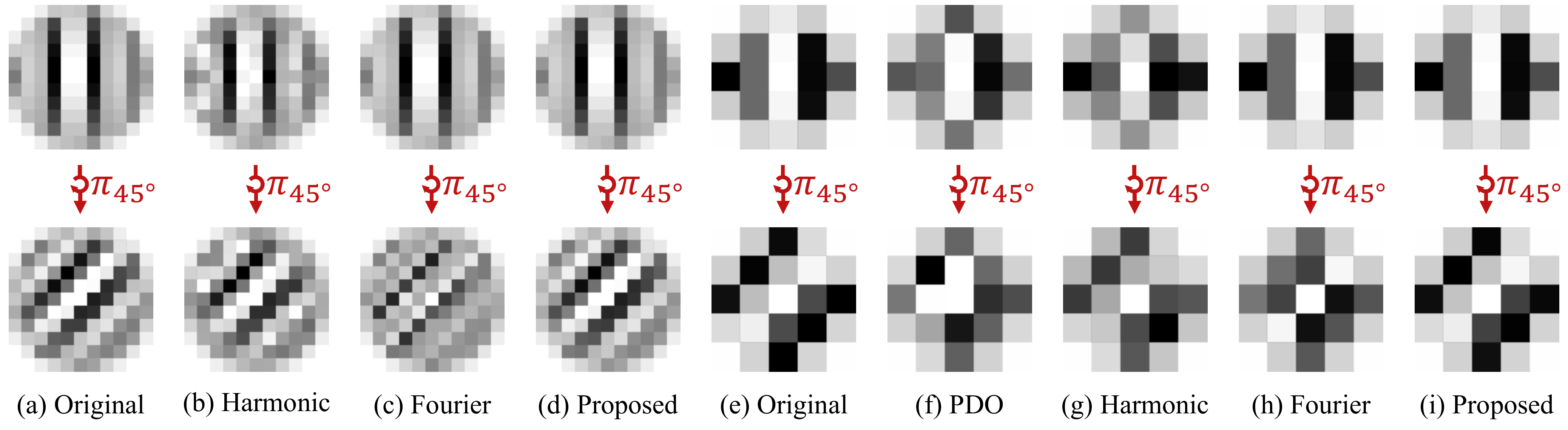}
\vspace{-8mm}
  \caption{(a) A discretization of the functional filter (\ref{Morlet}) and its $\nicefrac{\pi}{4}$ rotation, with filter size $p=11$ and mesh size $h= \nicefrac{1}{5}$.
   (b)-(d) The representations and the corresponding $\nicefrac{\pi}{4}$ rotations of the given 2D filter, where harmonics bases \cite{weiler2018learning},
    2D Fourier bases \cite{pei1998two} and the proposed bases in this study are adopted as basis functions, respectively.
   (e) A discretization of the functional filter (\ref{Morlet}), with filter size $p=5$ and mesh size $h= \nicefrac{1}{2}$.
   (f)-(i) The representations and the corresponding $\nicefrac{\pi}{4}$ rotations of the given 2D filter, by adopting PDO bases\cite{shen2020pdo},  harmonics bases \cite{weiler2018learning},
    2D Fourier bases \cite{pei1998two} and the proposed bases in this study as basis functions, respectively.}
\label{Fig:ConFilter}
\vspace{-1mm}
\end{figure*}

\textbf{Initialization scheme.}
%The weights $\hat{w}$   rather than pixel values.
An important practical issue of training deep networks is an appropriate initialization of $\hat{w}$. For convolutional networks,
Glorot and Bengio \cite{pmlr-v9-glorot10a} and He et. al. \cite{He_2015_ICCV}  came up with
initialization schemes which are accepted as a standard for random weight initialization.
Following \cite{weiler2018learning} and \cite{shen2020pdo},  we adapt He's
weight initialization scheme to initialize $\hat{w}$. Since $\|\tilde{\Psi}\| = \|\hat{w}\|$, this implies a normalization to the filter energies.

\section{Experimental Results}
In this section, we first conduct simulated experiments to evaluate the  representation
capability of the proposed filter parametrization regime. Experimental results on  classification and super-resolution tasks are then demonstrated to verify the effectiveness of the F-Conv method \footnote{Code is available at \url{https://github.com/XieQi2015/F-Conv}.}.

\subsection{Filter Parametrization Verification}
In Fig. \ref{Fig:Filter}, we have shown the superiority of the proposed bases when representing a simple filter function with rotation. Moreover, in Fig. \ref{Fig:Method}, we have further depicted the advantage of the proposed bases as compared to the corresponding traditional 2D Fourier bases after imposing a rotation on them.
In this section, we will show more verifications to evaluate the proposed filter parametrization method, compared with previous filter parametrization strategies, which are based on  traditional 2D Fourier bases (Fourier),  the harmonics (harmonic) \cite{weiler2018learning, weiler2019general} and partial differential operator (PDO)  \cite{shen2020pdo}, on more complex filters under more filter sizes. The latter two are actually the filter parametrization strategies exploited in SOTA equivariant convolutions, i.e., E2-CNN \cite{ weiler2019general} and PDO-eConv\cite{shen2020pdo}, respectively.

We estimate the representation coefficients by solving a least squares problem for all competing methods\footnote{For the 2D Fourier bases and the proposed bases, we can also exploit the 2D DFT to estimate the representation cofficients.}.  For example, for the general parametrization formulation (\ref{Parametrization}), we estimate $w_n$ by solving
\begin{equation}\label{w-solver}
  \min_{w_n} \sum_{i=1}^{p}\sum_{j=1}^p  \left(\hat{\psi}_{i,j}-\sum_{n-1}^Nw_n\psi_n\left(x_{ij}\right)\right)^2+\lambda \sum_{n=1}^N w_n^2,
\end{equation}
where $\hat{\psi}$ is the observed discrete filter for parametrization, $x_{ij}= \left(\left(i-\frac{p+1}{2}\right)h, \left(j-\frac{p+1}{2}\right)h\right)^T$ and we set  $\lambda=10^{-10}$. It is easy to obtain the closed-form solution for this optimization problem \cite{Boyd2004Convex}.

%\begin{figure}[t]
%\vspace{0mm}
%\hspace{-0mm}\includegraphics[width=1.0\linewidth]{ExpFig2_1.pdf}
%\vspace{-5mm}
%  \caption{(a) A discrete random filter generated by (\ref{Random}), with filter size $p=11$.
%   (b)-(d) The representations and correlated $\nicefrac{\pi}{4}$ rotations of a given 2D filter, where the we adopting harmonics basis \cite{weiler2018learning},
%    2D Fourier bases \cite{pei1998two} and the proposed bases as basis functions, respectively.}
%\label{Fig:ConFilter}
%\vspace{-0mm}
%\end{figure}
%
%\begin{figure}[t]
%\vspace{0mm}
%\hspace{-0mm}\includegraphics[width=1.0\linewidth]{ExpFig2_2.pdf}
%\vspace{-5mm}
%  \caption{(a) A discrete random filter generated by (\ref{Random}), with filter size $p=5$.
%   (b)-(e) The representations and correlated $\nicefrac{\pi}{4}$ rotations of a given 2D filter, by adopting PDO basis\cite{shen2020pdo},  harmonics basis \cite{weiler2018learning},
%    2D Fourier bases \cite{pei1998two} and the proposed bases as basis functions, respectively.}
%\label{Fig:ConFilter2}
%\vspace{-0mm}
%\end{figure}

\begin{table*}
  \caption{The RMSE (mean$\pm$standard deviation over $1000$ random generated samples) of filter parametrization on continuous functions, obtained by all competing methods under different filter sizes.}\vspace{-2mm}
  \label{ParametrizationExam}
  \centering  \setlength{\tabcolsep}{11pt}
  %\small
  \begin{tabular}{lcccccc}
    \toprule
             &\multicolumn{3}{c}{$11 \times 11$}   &\multicolumn{3}{c}{$5 \times 5$}                  \\
    \cmidrule(r){2-4}\cmidrule(r){5-7}
    Method   &\multicolumn{2}{c}{parameterization before rotation}   &   {simultaneously}    &\multicolumn{2}{c}{parameterization before rotation}   &   {simultaneously}    \\
        \cmidrule(r){2-3}\cmidrule(r){4-4}\cmidrule(r){5-6}\cmidrule(r){7-7}
               &    original        &$45^{\circ}$ rotation&   $ {0^{\circ} \& 45^{\circ}}$        &    original        &$45^{\circ}$ rotation&   $ {0^{\circ} \& 45^{\circ}}$ \\
    \midrule
        PDO    &    -                     &   -                     &    {-}                     &      4.3e-01$\pm$6.5e-02   &      1.1e-01$\pm$1.1e-01   &       {4.7e-01$\pm$3.8e-02} \\
   Harmonic    &    3.5e-01$\pm$4.3e-02   &   3.7e-01$\pm$4.5e-02   &    {2.7e-01$\pm$3.4e-02}   &      4.5e-01$\pm$6.5e-02   &      1.3e-01$\pm$1.3e-01   &       {3.3e-01$\pm$3.4e-02} \\
%  Harmonic+    &    1.3e-01$\pm$3.4e-02   &   1.6e-01$\pm$2.4e-02   &   1.2e-01$\pm$2.5e-02   &      6.2e-02$\pm$1.5e-02   &      6.4e-02$\pm$6.4e-02   &      5.4e-01$\pm$2.5e-02 \\
    Fourier    &    \textbf{9.7e-13$\pm$4.2e-14}   &   9.0e-01$\pm$2.4e-01   &    {3.7e-01$\pm$2.2e-01}   &      \textbf{1.7e-10$\pm$3.0e-11}   &      2.1e-01$\pm$2.1e-01   &       {1.8e-01$\pm$2.2e-01} \\
%   Fourier+    &    4.3e-01$\pm$4.2e-01   &   4.5e-01$\pm$4.2e-01   &   3.7e-01$\pm$3.5e-01   &      1.5e-01$\pm$1.3e-01   &      4.4e-01$\pm$4.4e-01   &      2.6e-01$\pm$3.5e-01 \\
%  Reviewers    &    9.6e-13$\pm$1.5e-13   &   4.1e-02$\pm$9.8e-03   &   2.0e-02$\pm$2.6e-03   &      1.7e-10$\pm$4.4e-11   &      2.4e-02$\pm$2.4e-02   &      6.6e-02$\pm$2.6e-03 \\
   Proposed    &    \textbf{9.7e-13$\pm$4.2e-14}   &   \textbf{4.1e-02$\pm$9.7e-03}   &    {\textbf{2.0e-02$\pm$2.6e-03}}   &      \textbf{1.7e-10$\pm$3.0e-11}   &      \textbf{2.4e-02$\pm$2.4e-02}   &      {\textbf{6.6e-02$\pm$2.6e-03}} \\
    \bottomrule
  \end{tabular}
  \vspace{-1.5mm}
\end{table*}

\textbf{Parametrization of continuous function.}
We first verify the representation capability of all compared methods on a continuous function. The exploited function is a 2D variant of Morlet wavelet \cite{Shyh1999Morlet}, i.e.,
\begin{equation}\label{Morlet}
  \psi_0(x) =  \exp\left({-\frac{1}{2}\left\|a\odot (x\!+\!b)\right\| ^2}\right)\cos\left(10\left(x_1\!+\!b_1\right)\right),
\end{equation}
where $\odot$ denotes the elemental wise multiplication, $a$ and $b$  are zooming and translation parameters, respectively. Let $\hat{\psi}_{ij} = \psi_0\left(x_{ij}\right)$, and then we solve $w_n$ for all competing methods by Eq. (\ref{w-solver}).

Fig. \ref{Fig:ConFilter} shows the representation results of (\ref{Morlet}) and the $45^{\circ}$ rotation results achieved by the representations of all competing methods, where $a = [2,1.5]^T$ and $b = [0.1,0.1]^T$. {Note that filter parametrization and filter rotation are perform separately, in order to show the degree of  aliasing effect.}
For the case of $p = 11$, the PDO based method is not compared since it is only designed for $5\times5$ filters. It is easy to observe that in this experiment, the results of the proposed method evidently outperform other comparison methods, with almost no difference from the ground truth. In comparison,  there is obvious quality degradation  in the parametrization result and its rotation for Harmonic based method. The commonly used Fourier bases can achieve an accurate parametrization result in rotation free case (similar as the proposed method). Nevertheless, its rotation result is with poor visual quality. For the comparison to PDO based filter parametrization, we also show the result of $p=5$ case in Fig. \ref{Fig:ConFilter} (e)-(i). It is seen that the result obtained by the proposed method perfectly preserves the shapes of the original function. Comparatively, the results by other competing methods contain more evident quality degradations.

 {
Table \ref{ParametrizationExam} shows more parametrization results on continuous functions, with $a = [2,1.5]^T$, $b\sim N(0,0,1)$ and $\theta$ be randomly sampled from $[0,2\pi)$. We perform filter parametrization on filter $\psi_{\theta} = \pi_{\theta}[\psi_0]$, where $\pi_{\theta}$ is defined in Eq. (\ref{Rotation}). We repeat experiments for 1000 times and calculate the  relative-mean-square error (RMSE) of parametrization in two cases. In the first case, we first perform  parametrization before rotation, and then calculate the RMSE for the original filter and its $45^{\circ}$ rotation separately. In the other case, we  perform  parametrization for a filter and its rotation simultaneously (sharing $w_n$), which is similar to the scene when filter parametrization is adopted in equivariant convolutions.
From Table \ref{ParametrizationExam}, it is easy to see that in the rotation free case, the 2D  Fourier bases and the proposed bases can represent the filter with RMSE lower than  $10^{-13}$, where the approximation error is possibly due to the numerical error in PC computing. When there is $45^{\circ}$ rotation, the Fourier bases parametrization performs relatively worse, the proposed method still evidently outperforms other competing methods. Besides, for the simultaneously parametrization case, it is easy to observe that the proposed method outperforms other competing methods.}

It is should be noted that the quality degradation of local filters may not largely affect the performance in high-level tasks, like classification and segmentation, which mainly require relatively  coarse-scale transformation equivariance knowledge. However, for low-level problems, one has to consider to represent much finer-grained local image details in the pixel level with higher accuracy requirement. Therefore, the proposed F-Conv is expected to show more superiority, as compared with current filter parametrization based equivariant convolutions, in low-level tasks.

\begin{table}
  \caption{The RMSE (mean$\pm$standard deviation over $1000$ random generated samples) of filter parametrization of random initialization filters, obtained by all competing methods under different filter sizes.}\vspace{-2mm}
  \label{ParametrizationExam2}
  \centering  \setlength{\tabcolsep}{16pt}
  %\small
  \begin{tabular}{lcc}
    \toprule
       Method           &{$11 \times 11$}   &{$5 \times 5$}                  \\
    \midrule
        PDO    &               -            &      4.9e-01$\pm$1.3e-01 \\
   Harmonic    &      4.2e-01$\pm$5.9e-02   &      5.7e-01$\pm$1.2e-01 \\
%  Harmonic+    &      2.0e-01$\pm$3.8e-02   &      1.8e-01$\pm$5.9e-02 \\
    Fourier    &      \textbf{9.5e-13$\pm$3.8e-14}   &      \textbf{6.2e-11$\pm$2.3e-11} \\
%   Fourier+    &      5.0e-01$\pm$7.1e-02   &      2.0e-01$\pm$7.7e-02 \\
% Fourier-1D    &      9.3e-13$\pm$5.8e-14   &      8.6e-11$\pm$2.9e-11 \\
   Proposed    &      \textbf{9.5e-13$\pm$3.8e-14}   &      \textbf{6.2e-11$\pm$2.3e-11} \\
    \bottomrule
  \end{tabular}
  \vspace{-1.5mm}
\end{table}

\textbf{Parametrization of random initialization.} In practice, filters are usually randomly initialized by  independent and identically distributed distributions such as  Gaussian  and uniform distributions.
Therefore, the representation capability on random initialization is important for filter parametrization methods. Specifically, in rotation equivariant convolutions,  a filter parametrization method is expected to be able to well represent and rotate the random initializations. Here, we verify all competing methods on the following randomly generated filter:
\begin{equation}\label{Random}
  \hat{\psi} = \mbox{Resize}(\mbox{Gaussion}(8,8), p) ,
\end{equation}
where $\mbox{Gaussian}(n, n)$ is a $n\times n$ 2D patch sampled from Gaussian distribution whose mean and variance are $0$ and $1$, respectively, and $\mbox{Resize}(\cdot,p)$ is the operator of resizing a 2D patch to the size of $p\times p$ with bicubic interpolation \cite{Han2013Interpolation}.

We repeat the experiments for 1000 times and calculate the  relative-mean-square error (RMSE) of filter parametrization of each completing method. The results are shown in the last 2 columns of Table \ref{ParametrizationExam2}. It is easy to observe that the 2D  Fourier bases and proposed bases can both exactly represent the filter, and both evidently outperform the harmonic and PDO based filter parametrizations.

 {\subsection{Equivariance Verification}}

In Section 4, we have provided the  theoretical analysis of equivariance error associated with the proposed convolutions. Here, we further empirically investigate the equivariance error of the proposed convolutions.

We first compare the equivariance error of CNN, E2-CNN \cite{weiler2019general} and the proposed F-Conv, where E2-CNN is the SOTA equivariant convolutions based on Harmonic bases. The 3 utilized networks are both with 5 convolutional layers, and each convolutional layer consistently contains a convolution, a batch normalization and a ReLU operator. We consider the $p24$  group for  E2-CNN and F-Conv methods and set the channel number of all convolution layers in the 2 utilized equivariant networks to be 9 (each channel contains 24 sub-channels for 24 orientations). We set the output layer of F-Conv  the same as  E2CNN (i.e., group max-pooling) for fair comparison in this experiment.   Besides, we set the channel number of all convolution layers in  CNN to be $9\times 24$ so that the three networks take similar computing memory.

\begin{table}[t]
  \caption{Average equivariant error of feature maps between rotation free and rotated inputs on 100 image in the Div2k testing samples.} \vspace{-2mm}
  \label{EQmeasure}
  \centering \setlength{\tabcolsep}{5pt}
  \begin{tabular}{lcccc}
    \toprule
                    &   \multicolumn{2}{c}{ Randomly initialization }          &  \multicolumn{2}{c}{ {Trained network} }\\
                     \cmidrule(r){2-3}\cmidrule(r){4-5}
    {Method}                  &   RMSE $\downarrow$&      LEPN (\%) $\downarrow$    &   {RMSE $\downarrow$}&      {LEPN (\%) $\downarrow$}     \\
    \midrule
    CNN                                             & 1.43$\pm$0.12   &    0.58$\pm$0.08 & {0.40$\pm$0.17}   &    {0.17$\pm$0.06}  \\
    E2-CNN                                          & 0.30$\pm$0.06   &    0.05$\pm$0.01 & {0.18$\pm$0.04}   &    {0.05$\pm$0.02}  \\
    %E2-CNN-C                                        & 0.09$\pm$0.03   &    0.02$\pm$0.01 & 0.16$\pm$0.04   &    0.04$\pm$0.02  \\
    F-Conv                                          & 0.42$\pm$0.08   &    0.10$\pm$0.05 & {0.19$\pm$0.05}   &    {0.06$\pm$0.03}  \\
    %F-Conv                                          & 0.12$\pm$0.03   &    0.02$\pm$0.02 & 0.17$\pm$0.04   &    0.04$\pm$0.02  \\
    \bottomrule
  \end{tabular}\vspace{-1.5mm}
\end{table}

We exploit 100  images from the testing set of the DIV2K dataset \cite{timofte2017ntire} for conducting the experiments. For each image, we rotated its output through the three models to a random degree $\theta\in[-\pi, \pi]$, and then compared it with the network output of the $\theta$ degree rotation of the same image.
{Table \ref{EQmeasure} shows equivariance errors of the three networks, before and after training them as auto-encoders, where we exploit two quality indices for performance evaluation. The first is RMSE, and  the second is large-error-pixel number (LEPN) which is the number of pixels satisfying $\nicefrac{\left\|x_r-x_0\right\|}{\left\|x_0\right\|}>1$, where $x_0$ is a pixel of the output of original image and $x_r$ is a pixel of the rotated copies. From the table, we can see that the equivariance errors of both E2-CNN and F-Conv are evidently less than that of the CNN model,  both in the trained and untrained networks, while E2-CNN achieves slightly better rotation equivariance than F-Conv.} The excellent rotation equivariance of E2-CNN is actually resulted from the heavy band limiting of its  filter parametrization \cite{timofte2017ntire}, which is detrimental to the expression capability of filters. As comparison, the F-Conv  achieves comparable rotation equivariance to E2-CNN without band limiting, which implies that it is a good choice for performing rotation equivariant convolution in low-level vision tasks.

%Fig. \ref{Fig:RandomIni} illustrates the output of CNN, E2-CNN and F-Conv  with an cartoon input image. It is easy to observe that the F-Conv and E2-CNN evidently better maintain the symmetry of local features underlying the image as compared to CNN. The outputs of F-Conv and E2-CNN show more orderly structural patterns, which better match the characteristics of the input images, while the output of CNN seems too chaotic to finely deliver such meaningful structures. Comparatively, the output of F-Conv shows clearer structure details than E2-CNN. These results imply that F-Conv can be more suitable for low-level vision tasks.

\begin{figure}[t]
\vspace{-0mm}
\hspace{-0mm}\includegraphics[width=1\linewidth]{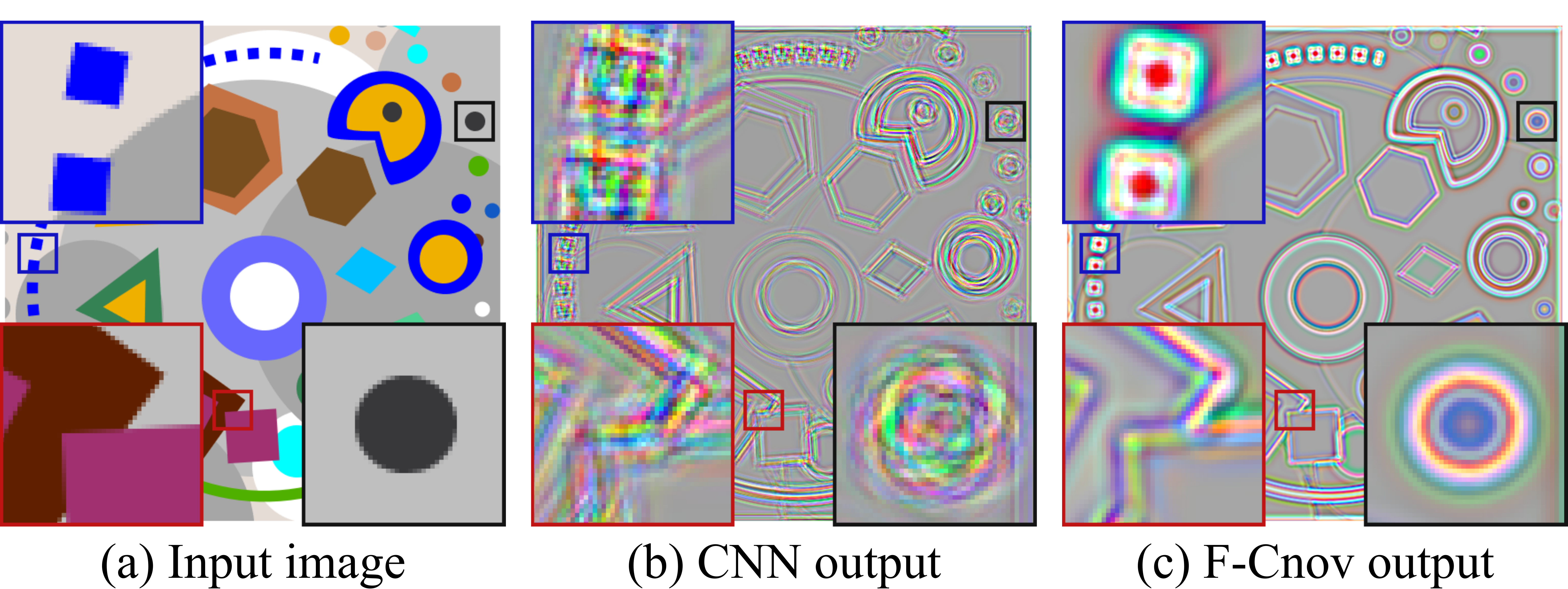}\centering
\vspace{-2mm}
  \caption{ {(a) A typical input cartoon image. (b)-(c) Outputs of randomly initialized CNN  and F-Conv, respectively, where the demarcated areas are zoomed in 5 times for easy observation.}
  }
\label{Fig:RandomIni}
\vspace{-1.5mm}
\end{figure}

Fig. \ref{Fig:RandomIni} illustrates the output of CNN  and F-Conv  with an cartoon input image. It is easy to observe that the outputs of F-Conv show more orderly structural patterns, which better match the characteristics of the input images. While the output of CNN seems too chaotic to finely deliver such meaningful structures.  These results verify the equivariance of F-Conv visually, and imply that F-Conv can be more suitable for low-level vision tasks than CNN.

\vspace{6mm}\subsection{Experimental Results on Image Classification}
MNIST-rot-12k \cite{Larochelle2007Empirical} is the standard benchmark for rotation-equivariant models.
It contains  handwritten digits of the classical
MNIST\footnote{\url{http://yann.lecun.com/exdb/mnist/}}, each being rotated by a random angle from $0$ to $2\pi$ (full angle).
The dataset contains 12,000 training samples and 50,000
test samples, respectively.
Meanwhile, we randomly select 2,000 training images as a validation set and choose the model with the lowest validation error during training.
Following \cite{weiler2018learning}, we augment the dataset with continuous rotations during the training progress.

%\textbf{Competing methods.} The competing methods including the following equivariant convolution methods:  G-CNN \cite{Cohen2016group}, H-Net \cite{Worrall2017Harmonic}, OR-TIPooling \cite{Zhou2017Oriented}, RotEqNet \cite{Marcos2017Rotation}, E2-CNN \cite{weiler2019general} and PDO-eConv \cite{shen2020pdo}. Note that the latest two method are the state of the art filter parametrization based equivariant convolution methods.

\textbf{Comparison in simple architectures.} We first evaluate the performance of F-Conv with comparison to the current equivariant
convolution methods. The competing methods include G-CNN \cite{Cohen2016group}, E2-CNN \cite{weiler2019general} and PDO-eConv \cite{shen2020pdo}, which are the most typical equivariant convolutions along this research line. Note that  the last two methods are the SOTA filter parametrization based equivariant convolution methods. We use the same network architectures with convolutions replaced by the equivariant convolutions proposed in all competing methods.

We consider the $p8$  group for  E2-CNN, PDO-eConv and the proposed F-Conv methods, and exploit a CNN model with  $6$ convolutional layers and a fully connected layer, while all the convolution layers and fully connected layer contain 10 channels. Each convolutional layer consistently contains a convolution, a batch normalization and a ReLU operator. We exploit $p4$ group for G-CNN since it is designed for $p4$ groups, and we set double times of the filter number for GCNN in each network layer to make its channel number similar with other methods. The filter sizes of these competing methods are all set as $5$.

 {The parameter number difference of these methods is mainly attributed to their different basis numbers for representing a filter (E2-CNN, PDO-eConv, F-Conv contain 11, 15,  25 bases, respectively). To observe the comparison under the same parameter number, we also conduct a F-Conv-reduce method which only utilizes the first 11 principal components of the proposed bases for filter parametrization. Note that F-Conv-reduce  contains exactly the same number of channels and parameters as E2-CNN method.}

We use Adam optimizer \cite{Diederik2015Adam}  with a weight decay of 0.001 to train the networks for 100 epoches. The batch size is set to be 128. The learning rate is started with $10^{-3}$ and reduced  gradually to $10^{-5}$.

 {Table \ref{Minist_Simple} shows the results of different methods, where all the experiments are averaged over 30 times repeating for guaranteeing the statistical stability of the results.  We can see that F-Conv-reduce achieves comparable performance as E2-CNN, while the complete F-Conv can achieve better performance with a little more computation cost.}

\textbf{Comparison with the leading board methods.} We then compare the performance of our F-Conv with  more competitive models. The competing methods include
H-Net \cite{Worrall2017Harmonic}, OR-TIPooling \cite{Zhou2017Oriented}, RotEqNet \cite{Marcos2017Rotation}, PTN-CNN\cite{esteves2017polar}, SFCNN \cite{weiler2018learning}, E2-CNN \cite{weiler2019general} and PDO-eConv \cite{shen2020pdo}.

\begin{table}
  \caption{Results of all competing methods with similar simple network architecture on MNIST-rot-12k.}\vspace{-2mm}
  \label{Minist_Simple}
  \centering   \setlength{\tabcolsep}{19pt}
  \begin{tabular}{lcc}
    \toprule
    Method                               &    Test Error (\%) &   Params \\
    \midrule
    G-CNN \cite{Cohen2016group}          &          2.00$\pm$0.106           & 201.7k\\
%    H-Net \cite{Worrall2017Harmonic}    &          1.57           & 99k  \\
    E2-CNN \cite{weiler2019general}      &          1.23$\pm$0.068          & 44.6k \\
    PDO-eConv \cite{shen2020pdo}         &          1.79$\pm$0.138          & 60.6k \\
     {F-Conv-reduce}                        &           {1.23$\pm$0.091}          &  {44.6k}           \\
%    F-Conv-1D                            &          1.27$\pm$0.082          & 100.9k           \\
    F-Conv                               &          \textbf{1.13$\pm$0.059} & 100.9k           \\
    \bottomrule
  \end{tabular}\vspace{-1.5mm}
\end{table}

Following  E2-CNN \cite{weiler2019general}, we set the group size and filter size  to be $12$ and $5\time 5$, respectively. The model is trained using the Adam algorithm \cite{Diederik2015Adam}  with a weight decay of 0.001. We train the network with the batch size 128 for 200 epochs. The initial learning rate is set as 5e-3 and reduced  gradually to 5e-6, and the dropout rate is 0.25.
We exploit a CNN model with  $7$ convolutional layers and a fully connected network. The convolutional layers have  $16$, $16$, $32$, $32$, $32$, $64$ and $96$ channels, respectively. Each convolutional layer of F-Conv consistently includes a batch normalization, a ReLU  and a dropout operator.  We use spatial pooling and orientation pooling after the final F-Conv layer, and input the pooling result to the fully connected network, containing 1 hidden layer, with hidden node number set as $96$.

Table \ref{Minist_SOTA} shows the results of F-Conv on MNIST-rot-12k, as well as the reported best results of leading board  methods on this dataset\footnote{Since the parameter number of OR-TIPooling, SFCNN and E2-CNN are not reported in their paper, we thus calculate the approximate ones according the reported networks.}. As shown in the table,  the SOTA methods have achieved high precisions (e.g., $0.682\%$,  $0.714\%$ and $0.709\%$ test error for E2-CNN, SFCNN and PDO-eConv, respectively). Comparatively, our method achieves $0.671\%$ test error, getting further performance gain.
This substantiates the effectiveness of F-Conv in this high-level task.

\begin{table}
  \caption{Results of leading board  methods and ours on MNIST-rot-12k.} \vspace{-2mm}
  \label{Minist_SOTA}
  \centering \setlength{\tabcolsep}{18pt}
   {\begin{tabular}{llc}
    \toprule
    {Method}         &    Test Error (\%)   &   Params \\
    \midrule
%    G-CNN \cite{Cohen2016group}                     &          2.74             \\
    H-Net \cite{Worrall2017Harmonic}                &          1.69                     & 0.03M \\
    OR-TIPooling \cite{Zhou2017Oriented}            &          1.54                     & $\approx$1M\\
    RotEqNet \cite{Marcos2017Rotation}              &          1.01                     & 0.10M \\
    PTN-CNN\cite{esteves2017polar}                  &          0.89                     & 0.25M \\
    SFCNN \cite{weiler2018learning}                 &          0.714$\pm$0.022          &  $\approx$3M \\ %2.68M
    E2-CNN \cite{weiler2019general}                 &          0.682$\pm$0.022          &  $\approx$5M \\ %5.35M
    PDO-eConv \cite{shen2020pdo}                    &          0.709                    & 0.65M \\
%    F-Conv-rev                                      &          0.684$\pm$0.028          & 3.05M \\
    F-Conv                                          &         \textbf{0.671$\pm$0.020}  & 3.05M \\
    \bottomrule
  \end{tabular}}\vspace{-1mm}
\end{table}

\begin{table*}
  \caption{The average SR results {(mean $\!\!\pm\!$ deviation)} of all competing methods on 4 exploited image datasets, including Urban100 \cite{huang2015single}, B100 \cite{martin2001database}, Set14 \cite{zeyde2010single} and Set5 \cite{bevilacqua2012low}. The results in the upper and lower parts are produced by networks trained without and with data argumentation, respectively.}\vspace{-1mm}
  \label{SR_MeanResult}
  \centering   \setlength{\tabcolsep}{1.7pt}
   {\begin{tabular}{lccccccccccccc}
    \toprule
                    &&    \multicolumn{6}{c}{$\times 2$}  &    \multicolumn{6}{c}{$\times 4$}  \\
    \cmidrule(r){3-8}\cmidrule(r){9-14}
     {Method}       & argu.  &\multicolumn{2}{c}{EDSR \cite{lim2017enhanced}}&\multicolumn{2}{c}{RDN \cite{zhang2018residual}}&\multicolumn{2}{c}{RCAN \cite{zhang2018image}}&
     \multicolumn{2}{c}{EDSR \cite{lim2017enhanced}}&\multicolumn{2}{c}{RDN \cite{zhang2018residual}}&\multicolumn{2}{c}{RCAN \cite{zhang2018image}}\\
    \cmidrule(r){3-4}\cmidrule(r){5-6}\cmidrule(r){7-8}\cmidrule(r){9-10}\cmidrule(r){11-12}\cmidrule(r){13-14}
                    &&PSNR&SSIM&PSNR&SSIM&PSNR&SSIM&PSNR&SSIM&PSNR&SSIM&PSNR&SSIM\\
    \midrule
                  CNN & $\bm{\times}$& 32.39 {\tiny $\!\!\pm\!$ 4.81} & 0.917 {\tiny $\!\!\pm\!$ 0.053} & 32.38 {\tiny $\!\!\pm\!$ 4.83} & 0.917 {\tiny $\!\!\pm\!$ 0.053} & 32.48 {\tiny $\!\!\pm\!$ 4.85} & 0.919 {\tiny $\!\!\pm\!$ 0.052} & 26.95 {\tiny $\!\!\pm\!$ 4.15} & 0.768 {\tiny $\!\!\pm\!$ 0.109} & 27.05 {\tiny $\!\!\pm\!$ 4.21} & 0.771 {\tiny $\!\!\pm\!$ 0.108} & 27.02 {\tiny $\!\!\pm\!$ 4.19} & 0.770 {\tiny $\!\!\pm\!$ 0.109}  \\
               G-CNN  & $\bm{\times}$& 32.44 {\tiny $\!\!\pm\!$ 4.81} & \textbf{0.918 {\tiny $\!\!\pm\!$ 0.053}} & 32.45 {\tiny $\!\!\pm\!$ 4.89} & 0.918 {\tiny $\!\!\pm\!$ 0.053} & 32.57 {\tiny $\!\!\pm\!$ 4.84} & 0.919 {\tiny $\!\!\pm\!$ 0.052} & 27.02 {\tiny $\!\!\pm\!$ 4.09} & 0.770 {\tiny $\!\!\pm\!$ 0.107} & 27.08 {\tiny $\!\!\pm\!$ 4.12} & 0.772 {\tiny $\!\!\pm\!$ 0.107} & 27.13 {\tiny $\!\!\pm\!$ 4.16} & 0.774 {\tiny $\!\!\pm\!$ 0.107}  \\
              E2-CNN  & $\bm{\times}$& 32.23 {\tiny $\!\!\pm\!$ 4.72} & 0.916 {\tiny $\!\!\pm\!$ 0.053} & 32.33 {\tiny $\!\!\pm\!$ 4.82} & 0.917 {\tiny $\!\!\pm\!$ 0.053} & 32.01 {\tiny $\!\!\pm\!$ 4.67} & 0.914 {\tiny $\!\!\pm\!$ 0.055} & 26.95 {\tiny $\!\!\pm\!$ 4.06} & 0.768 {\tiny $\!\!\pm\!$ 0.108} & 27.03 {\tiny $\!\!\pm\!$ 4.08} & 0.769 {\tiny $\!\!\pm\!$ 0.108} & 26.72 {\tiny $\!\!\pm\!$ 3.87} & 0.760 {\tiny $\!\!\pm\!$ 0.109}  \\
            PDO-eConv & $\bm{\times}$& 31.62 {\tiny $\!\!\pm\!$ 4.65} & 0.909 {\tiny $\!\!\pm\!$ 0.057} & 31.41 {\tiny $\!\!\pm\!$ 4.60} & 0.906 {\tiny $\!\!\pm\!$ 0.058} & 31.79 {\tiny $\!\!\pm\!$ 4.70} & 0.911 {\tiny $\!\!\pm\!$ 0.056} & 26.56 {\tiny $\!\!\pm\!$ 3.89} & 0.752 {\tiny $\!\!\pm\!$ 0.113} & 25.77 {\tiny $\!\!\pm\!$ 3.44} & 0.726 {\tiny $\!\!\pm\!$ 0.111} & 26.55 {\tiny $\!\!\pm\!$ 3.83} & 0.753 {\tiny $\!\!\pm\!$ 0.111}  \\
               F-Conv & $\bm{\times}$& \textbf{32.51 {\tiny $\!\!\pm\!$ 4.81}} & \textbf{0.918 {\tiny $\!\!\pm\!$ 0.053}} & \textbf{32.62 {\tiny $\!\!\pm\!$ 4.87}} & \textbf{0.920 {\tiny $\!\!\pm\!$ 0.052}} & \textbf{32.73 {\tiny $\!\!\pm\!$ 4.83}} & \textbf{0.920 {\tiny $\!\!\pm\!$ 0.051}} & \textbf{27.15 {\tiny $\!\!\pm\!$ 4.19}} & \textbf{0.774 {\tiny $\!\!\pm\!$ 0.107}} & \textbf{27.21 {\tiny $\!\!\pm\!$ 4.22}} & \textbf{0.776 {\tiny $\!\!\pm\!$ 0.106}} & \textbf{27.20 {\tiny $\!\!\pm\!$ 4.16}} & \textbf{0.775 {\tiny $\!\!\pm\!$ 0.105}}  \\
 \midrule
                  CNN & $\checkmark$& 32.41 {\tiny $\!\!\pm\!$ 4.80} & 0.918 {\tiny $\!\!\pm\!$ 0.053} & 32.45 {\tiny $\!\!\pm\!$ 4.83} & 0.918 {\tiny $\!\!\pm\!$ 0.053} & 32.52 {\tiny $\!\!\pm\!$ 4.84} & 0.919 {\tiny $\!\!\pm\!$ 0.052} & 27.18 {\tiny $\!\!\pm\!$ 4.21} & 0.774 {\tiny $\!\!\pm\!$ 0.107} & 27.20 {\tiny $\!\!\pm\!$ 4.20} & 0.775 {\tiny $\!\!\pm\!$ 0.107} & 27.26 {\tiny $\!\!\pm\!$ 4.19} & 0.776 {\tiny $\!\!\pm\!$ 0.106}  \\
               G-CNN  & $\checkmark$& 32.46 {\tiny $\!\!\pm\!$ 4.76} & 0.918 {\tiny $\!\!\pm\!$ 0.052} & 32.45 {\tiny $\!\!\pm\!$ 4.85} & 0.918 {\tiny $\!\!\pm\!$ 0.053} & 32.54 {\tiny $\!\!\pm\!$ 4.80} & 0.919 {\tiny $\!\!\pm\!$ 0.051} & 27.18 {\tiny $\!\!\pm\!$ 4.15} & 0.775 {\tiny $\!\!\pm\!$ 0.106} & 27.12 {\tiny $\!\!\pm\!$ 4.11} & 0.773 {\tiny $\!\!\pm\!$ 0.106} & 27.21 {\tiny $\!\!\pm\!$ 4.23} & 0.775 {\tiny $\!\!\pm\!$ 0.106}  \\
              E2-CNN  & $\checkmark$& 32.30 {\tiny $\!\!\pm\!$ 4.74} & 0.917 {\tiny $\!\!\pm\!$ 0.053} & 32.36 {\tiny $\!\!\pm\!$ 4.85} & 0.917 {\tiny $\!\!\pm\!$ 0.053} & 32.03 {\tiny $\!\!\pm\!$ 4.69} & 0.914 {\tiny $\!\!\pm\!$ 0.055} & 26.95 {\tiny $\!\!\pm\!$ 4.09} & 0.767 {\tiny $\!\!\pm\!$ 0.109} & 27.06 {\tiny $\!\!\pm\!$ 4.10} & 0.771 {\tiny $\!\!\pm\!$ 0.107} & 26.77 {\tiny $\!\!\pm\!$ 3.90} & 0.761 {\tiny $\!\!\pm\!$ 0.109}  \\
            PDO-eConv & $\checkmark$& 31.65 {\tiny $\!\!\pm\!$ 4.68} & 0.909 {\tiny $\!\!\pm\!$ 0.057} & 30.76 {\tiny $\!\!\pm\!$ 4.24} & 0.899 {\tiny $\!\!\pm\!$ 0.059} & 31.75 {\tiny $\!\!\pm\!$ 4.71} & 0.910 {\tiny $\!\!\pm\!$ 0.057} & 26.59 {\tiny $\!\!\pm\!$ 3.91} & 0.754 {\tiny $\!\!\pm\!$ 0.113} & 25.83 {\tiny $\!\!\pm\!$ 3.48} & 0.728 {\tiny $\!\!\pm\!$ 0.111} & 26.51 {\tiny $\!\!\pm\!$ 3.79} & 0.752 {\tiny $\!\!\pm\!$ 0.111}  \\
               F-Conv & $\checkmark$& \textbf{32.52 {\tiny $\!\!\pm\!$ 4.81}} & \textbf{0.919 {\tiny $\!\!\pm\!$ 0.053}} & \textbf{32.66 {\tiny $\!\!\pm\!$ 4.82}} & \textbf{0.920 {\tiny $\!\!\pm\!$ 0.051}} & \textbf{32.73 {\tiny $\!\!\pm\!$ 4.88}} & \textbf{0.921 {\tiny $\!\!\pm\!$ 0.050}} & \textbf{27.25 {\tiny $\!\!\pm\!$ 4.23}} & \textbf{0.776 {\tiny $\!\!\pm\!$ 0.106}} & \textbf{27.26 {\tiny $\!\!\pm\!$ 4.20}} & \textbf{0.777 {\tiny $\!\!\pm\!$ 0.106}} & \textbf{27.32 {\tiny $\!\!\pm\!$ 4.21}} & \textbf{0.778 {\tiny $\!\!\pm\!$ 0.104}}  \\

\bottomrule
  \end{tabular}}\vspace{-0mm}
\end{table*}

\begin{figure*}[t]
\vspace{-0mm}
\hspace{-0mm}\includegraphics[width=1.0\linewidth]{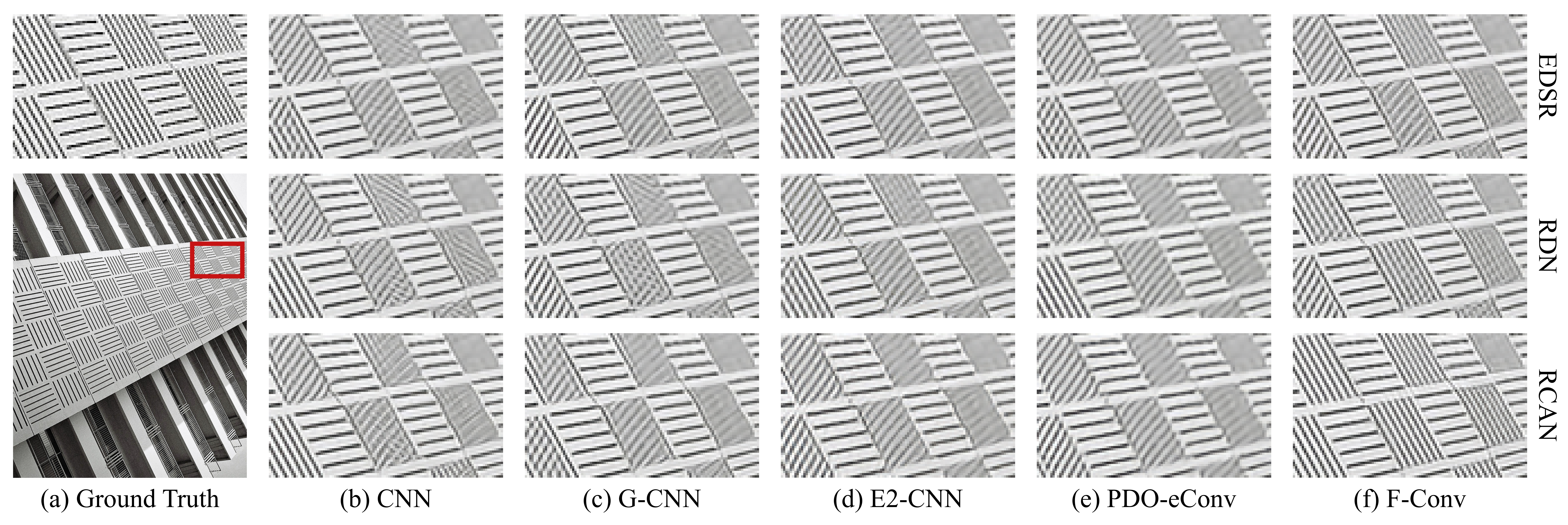}
\vspace{-7.5mm}
  \caption{(a)  A sample of high-resolution image from the Urban100 \cite{huang2015single} dataset. (b) From upper to lower: the 2 times super-resolution images restored by the EDSR, RDN and RCAN methods, respectively, where the convolution operators are set as commonly used convolutions, i.e., CNN. (c)-(f) From upper to lower: the super-resolution images restored by the EDSR, RDN and RCAN methods, respectively, where the convolution operators are set as G-CNN, E2-CNN, PDO-eConv and the proposed F-Conv, respectively. All the involved methods are trained without data argumentation. }
\label{Fig:SR_X2NoArgument}
\vspace{-1.5mm}
\end{figure*}

%\begin{figure*}[t]
%\vspace{0mm}
%\hspace{-0mm}\includegraphics[width=1.0\linewidth]{ExpFigSR1x4.pdf}
%\vspace{-8mm}
%  \caption{(a)  A sample of high-resolution image from the Urban100 \cite{huang2015single} dataset. (b) From upper to lower: the 4 times super-resolution images restored by the EDSR, RDN and RCAN methods, respectively, where the convolution operators are set as commonly used convolutions, i.e., CNN. (c)-(f) From upper to lower: the super-resolution images restored by the EDSR, RDN and RCAN methods, respectively, where the convolution operators are set as G-CNN, E2-CNN, PDO-eConv and the proposed F-Conv, respectively. All the involved methods are trained without data argumentation. }
%\label{Fig:SR_X4NoArgument}
%\vspace{-1.5mm}
%\end{figure*}
%
%\begin{figure*}[t]
%\vspace{0mm}
%\hspace{-0mm}\includegraphics[width=1.0\linewidth]{ExpFigSR1_argument_X2.pdf}
%\vspace{-8mm}
%  \caption{(a)  A sample of high-resolution image from the Urban100 \cite{huang2015single} dataset.
%  (b) From upper to lower: the 2 times super-resolution images restored by the EDSR, RDN and RCAN methods, respectively, where the convolution operators are set as commonly used convolutions, i.e., CNN. (c)-(f) From upper to lower: the super-resolution images restored by the EDSR, RDN and RCAN methods, respectively, where the convolution operators are set as G-CNN, E2-CNN, PDO-eConv and the proposed F-Conv, respectively. All the involved methods are trained with data argumentation. }
%\label{Fig:SR_X2}
%\vspace{-2mm}
%\end{figure*}

\begin{figure*}[t]
\vspace{0.5mm}
\hspace{-0mm}\includegraphics[width=1.0\linewidth]{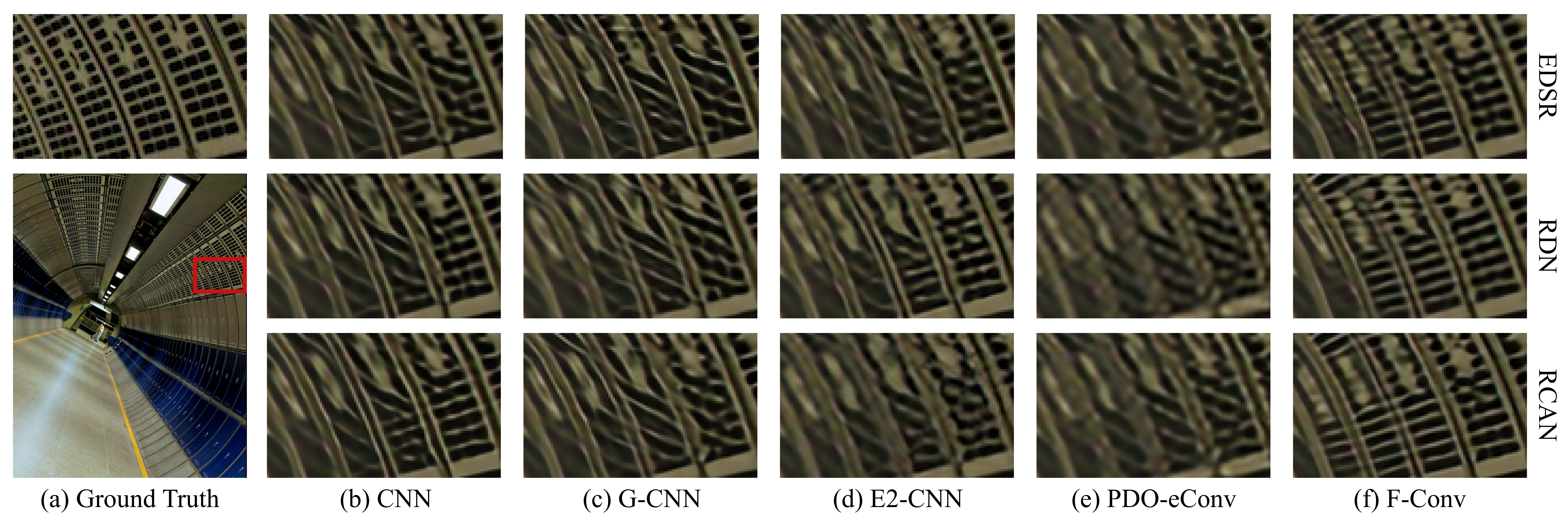}
\vspace{-7.5mm}
  \caption{(a)  A sample of high resolution image from the Urban100 \cite{huang2015single} dataset.
  (b) From upper to lower: the 4 times super-resolution images restored by the EDSR, RDN and RCAN methods, respectively, where the convolution operators are set as commonly used convolutions, i.e., CNN. (c)-(f) From upper to lower: the super-resolution images restored by the EDSR, RDN and RCAN methods, respectively, where the convolution operators are set as G-CNN, E2-CNN, PDO-eConv and the proposed F-Conv, respectively. All the involved methods are trained with data argumentation. }
\label{Fig:SR_X4}
\vspace{-1.5mm}
\end{figure*}

\subsection{Experimental Results on Image Super-resolution}

We then adopt the filter parameterized equivariant convolutions to a typical low-level image processing task, image super-resolution, for verifying their performance in low-level problems. Specifically, single image super-resolution (SR), as a classical problem in computer vision, aims at recovering a high-resolution image from a single low-resolution image.
%This problem is inherently ill-posed since a multiplicity of solutions exist for any given low-resolution pixel.
Recently, deep convolutional neural network (CNN) based methods have achieved significant improvements over conventional SR methods \cite{wang2020deep,dong2014learning,lim2017enhanced,zhang2018residual,zhang2018image}, and attracted much research attention.

\textbf{Network architecture settings.}
We exploit 3 SOTA networks designs for SR tasks, including EDSR \cite{lim2017enhanced}, RDN \cite{zhang2018residual} and RCAN \cite{zhang2018image}, for our experiments.
The EDSR network consists of residual blocks \cite{he2016deep}, a global residual connection and an upsampling modular. In our experiments,  we set the number of residual blocks  to be $16$, and set the channel number of each residual block to be $256$.  The RDN network consists of residual blocks  \cite{he2016deep}, dense connections \cite{huang2017densely} and an upsampling modular. We set the number of residual blocks, the growth rate and conv number in RDN as $16$, $64$ and $8$, respectively.  The RCAN network consists of residual in residual (RIR) blocks and residual channel attention blocks. Wet set the number of RIR blocks, the number of residual blocks of RIR and the channel number of each residual block to be $10$ and $16$ and $64$, respectively.

\textbf{Convolutional method settings.}
The competing  equivariant convolution methods include G-CNN \cite{Cohen2016group}, E2-CNN \cite{weiler2019general}, PDO-eConv \cite{shen2020pdo}  and the proposed F-Conv. We replace the original convolutions in EDSR, RDN and RCAN with the competing convolutions, respectively.

For G-CNN based network, since an intermediate feature map is a $4$-channel matrix, we set the channel number of residual blocks as $\nicefrac{1}{4}$ of that in the original network, so as to make it with similar memory cost as the original one. Following the original setting in EDSR, RDN and RCAN, the filter size is set as $3\times 3$ in G-CNN based network. For E2-CNN, PDO-eConv  and F-Conv,  we  exploit $p8$ group for the equivariant convolutions and set the channel number of each residual block (and the growth rate in RDN) as $\nicefrac{1}{8}$ to the original networks in each layer to keep their similar memory with the original network. Besides, we set the filter size as $5\times 5$ for these convolutions (this is due to that the PDO-eCOnv is only designed for filters with size $5\times 5$, and the circular shape masks in filters of E2-CNN and F-Conv make $3\times 3$ filters contain insufficient effective pixels). Since the channel attention operation is not rotation equivariant, we remove the  channel attention modular in the equivariant-convolution-based RCAN methods. For these equivariant based methods, we adopt the proposed convolution methods for output layer before the upsampling operator, which ensures that the major parts of these networks are rotation equivariant. It should be indicated that the upsampling operator at the tail parts are still not rotation equivariant, which does not substantially influence the effect brought by these rotation equivariant convolutions.

\textbf{Datasets and degradation models.}
Following  \cite{lim2017enhanced, zhang2018residual,zhang2018image}, we use 800 training images from the DIV2K dataset \cite{timofte2017ntire} as the training set. For testing, we use four standard benchmark datasets,  including Urban100 \cite{huang2015single}, B100 \cite{martin2001database}, Set14 \cite{zeyde2010single} and Set5 \cite{bevilacqua2012low}, which contain 100, 100, 14 and 5 natural images, respectively. We conduct experiments with bicubic degradation models  \cite{lim2017enhanced, zhang2018residual,zhang2018image}.

\textbf{Training settings.} For all competing methods in this series of experiments, we use Adam optimizer \cite{Diederik2015Adam} with no weight decay to train the networks for 150 epoches.
Besides, following the setting of the original paper of EDSR \cite{lim2017enhanced}, RDN \cite{zhang2018residual} and RCAN \cite{zhang2018image}, during the training, we set the batch size for all the EDSR, RDN and RCAN based methods as 16. For the 2 scale super-resolution cases, we set the training patch size for EDSR, RDN and RCAN based methods as 96, 64 and 96,  respectively, and set the training patch size for the 4 scale super-resolution cases double to the  2 scale cases.  The initial leaning rate is set as $2\times10^{-4}$ and then decreased to half at the  $100$ and $130$ epoches.

Since the rotation-based data augmentation will lead to more ``rotation equivariant" training results, and this effect may be mixed with the benefit resulted from rotation equivariant convolutions, we first train the competing methods without data augmentation, in order to observe the pure benefit from equivariant convolutions. Then, following the setting of previous methods, we also train the competing methods by randomly rotating the training images  by  $90^{\circ}$, $180^{\circ}$, $270^{\circ}$ and flipping horizontally for data augmentation.

\textbf{Quantitative results.} Table \ref{SR_MeanResult} shows the SR results of 15 competing methods  on 5 exploited data sets, without and with data augmentation, respectively.  The evaluation measures include PSNR and SSIM \cite{wang2004image} on Y channel (i.e., luminance) of the transformed YCbCr space. By comparing the results in Table \ref{SR_MeanResult}, it is easy to see that removing data augmentation will significantly degenerate the performance of the 3 CNN-based SR methods, especially in the 4 scale SR experiments. Comparatively, the performance of equivariant-convolution-based methods does not change too much with or without data augmentation, since most of the network modulars in these methods are rotation equivariant. This phenomena imply that the rotation symmetries of local features and rotation equivariant convolutions should be helpful for improving SR performance. Besides, one can also observe that E2-CNN and PDO-eConv have not attained satisfactory SR performance, possibly due to their less representation accuracy of the filter parametrization methods. Comparatively, the proposed F-Conv method shows evident superiority than other filter-parametrization-based equivariant convolutions in this task, and achieves better result than CNN and G-CNN based methods which do not involve filter parametrization. To the best of our knowledge, this should be the first filter parametrization based rotation equivariant convolutions that help intrinsically improve performance of low-level tasks.

\textbf{Visual results.} Figs. \ref{Fig:SR_X2NoArgument} and \ref{Fig:SR_X4} visually show the SR results of all competing methods on 2 samples from the Unban100 dataset, with SR scale set as 2, trained without augmentation,  and SR scale set as 4, trained with data augmentation, respectively. From these figures, it is easy to observe that the SR results of filter parametrization based method like E2-CNN and PDO-eConv based methods are usually with over-smooth artifacts and certainly lose texture details. As comparison, F-Conv based method performs evidently better in sense of achieving clearer SR image and more faithfully preserving image textures and edges, which are superior to the methods based on other 4 convolutions.
This implies that the proposed filter parametrization method should be helpful to enhance the performance on a low-level task.

%\begin{table*}
%  \caption{ {Maybe we can use this table }}
%  \label{SR_MeanResult_mean}
%  \centering
%  \setlength{\tabcolsep}{10.5pt}
%  \begin{tabular}{lcccccccccc}
%    \toprule
%    \multirow{2}{*}{Method}       & \multirow{2}{*}{Scale}    &\multicolumn{3}{c}{EDSR \cite{lim2017enhanced}}&\multicolumn{3}{c}{RDN \cite{zhang2018residual}}&\multicolumn{3}{c}{RCAN \cite{zhang2018image}}\\
%                    \cmidrule(r){3-5} \cmidrule(r){6-8} \cmidrule(r){9-11}
%                    &&PNSR&SSIM&Params&PSNR&SSIM&Params&PSNR&SSIM&Params\\
%    \midrule
%    CNN                                 &$\times 2$& 31.228& 0.893& &31.782& 0.900& &31.796&0.900& \\
%    G-CNN \cite{Cohen2016group}         &$\times 2$& 31.018& 0.890& &31.760& 0.900& &29.540&0.867& \\
%    E2-CNN \cite{weiler2019general}     &$\times 2$& 31.076& 0.891& &31.326& 0.894& &29.562&0.896& \\
%    PDO-eConv \cite{shen2020pdo}        &$\times 2$& 30.485& 0.882& &31.007& 0.890& &29.554&0.868& \\
%    F-Conv                              &$\times 2$& 31.281& 0.893& &31.828& 0.901& &31.883&0.901& \\
%    \midrule
%    CNN                                 &$\times 4$& & & & & & & & & \\
%    G-CNN \cite{Cohen2016group}         &$\times 4$& & & & & & & & & \\
%    E2-CNN \cite{weiler2019general}     &$\times 4$& & & & & & & & & \\
%    PDO-eConv \cite{shen2020pdo}        &$\times 4$& & & & & & & & & \\
%    F-Conv                              &$\times 4$& & & & & & & & & \\
%    \bottomrule
%  \end{tabular}
%\end{table*}

\section{Conclusion}

In this paper, we have proposed a filter parametrization method, and built rotation equivariant convolutions basing on it. The proposed filter parametrization can be viewed as an enhanced version of conventional Fourier series expansion, where we reduce the maximum frequency of the bases by exploiting  symmetrical functions of low frequency Fourier bases to replace the high frequency ones.
By theoretical analysis and empirical evaluation, we have shown that the proposed filter parametrization method inherits the high-accuracy of Fourier series expansion for representing functional filter when there is no rotation, and more importantly, it significantly alleviates the heavy aliasing effect which original Fourier series expansion suffers from.
In this way, the proposed filter parametrization method evidently prompts the low-accuracy representation issue of previous filter parametrization methods.
Based on this filter parametrization manner, we construct a new equivariant convolution framework, named F-Conv, and analyze its theoretical properties in detail. We have further demonstrated the superiority of the proposed F-Conv  beyond previous filter parametrization methods. Especially, to the best of our knowledge, F-Conv should be the first equivariant convolution that could intrinsically help evidently improve the performance of the low-level task, implying its better preservation capability of the rotation symmetries (more than 4 angles) of image features, and potential usefulness for a wider range of image processing tasks.

Except rotation equivariant convolutions, the proposed Fourier series expansion based filter parametrization should be useful for the designing of many other learnable operators in deep learning frameworks.
In our future work,  we will further extend the employed filter parametrization methodology and explore more applications along this line. Typically, the multi-scale network modules play an important role in many computer vision tasks, and the utilized filter parametrization framework in this study is hopeful to be extended to design novel and rational multi-scale filters. A feasible strategy might be to discretize the functional filter with different resolutions, and then multi-scale filters with shared parameters can be obtained. Besides, the parametrization strategy can also be used in dynamic network design, where one can input different kinds of transformations to a learnable filter to dynamically control the network.
{A more comprehensive and deep exploration on filter parametrization method with high accuracy should also be a meaningful research issue in future research. Besides the Fourier series expansion, other transforms, such as Hartley Transform\cite{bracewell1983discrete} and Cosine Transform \cite{ahmed1974discrete}, can be also exploited for designing efficient and high accuracy filter parametrization methods.}

\section*{Acknowledgement}
We want to sincerely thank the anonymous reviewers for their constructive comments and suggestions, which have helped significantly improve the quality of this paper.

This research was supported by NSFC project under contracts U21A6005, 61721002, U1811461, 62076196, the Major Key Project of PCL under contract PCL2021A12, and the Macao Science and Technology Development Fund under Grant 061/2020/A2.

\bibliographystyle{unsrt}
%\bibliography{egbib}

\begin{thebibliography}{10}

\bibitem{zeiler2014visualizing}
Matthew~D Zeiler and Rob Fergus.
\newblock Visualizing and understanding convolutional networks.
\newblock In {\em European conference on computer vision}, pages 818--833.
  Springer, 2014.

\bibitem{szegedy2015going}
Christian Szegedy, Wei Liu, Yangqing Jia, Pierre Sermanet, Scott Reed, Dragomir
  Anguelov, Dumitru Erhan, Vincent Vanhoucke, and Andrew Rabinovich.
\newblock Going deeper with convolutions.
\newblock In {\em Proceedings of the IEEE conference on computer vision and
  pattern recognition}, pages 1--9, 2015.

\bibitem{weiler2018learning}
Maurice Weiler, Fred~A Hamprecht, and Martin Storath.
\newblock Learning steerable filters for rotation equivariant cnns.
\newblock In {\em Proceedings of the IEEE Conference on Computer Vision and
  Pattern Recognition}, pages 849--858, 2018.

\bibitem{weiler2019general}
Maurice Weiler and Gabriele Cesa.
\newblock General $ e (2) $-equivariant steerable cnns.
\newblock 2019.

\bibitem{shen2020pdo}
Zhengyang Shen, Lingshen He, Zhouchen Lin, and Jinwen Ma.
\newblock Pdo-econvs: Partial differential operator based equivariant
  convolutions.
\newblock In {\em International Conference on Machine Learning}, pages
  8697--8706. PMLR, 2020.

\bibitem{Cohen2016group}
Taco Cohen and Max Welling.
\newblock Group equivariant convolutional networks.
\newblock In {\em Proceedings of The 33rd International Conference on Machine
  Learning}, pages 2990--2999. PMLR, 2016.

\bibitem{hoogeboom2018hexaconv}
Emiel Hoogeboom, Jorn W.~T. Peters, Taco~S. Cohen, and Max Welling.
\newblock Hexaconv, 2018.

\bibitem{pei1998two}
Soo-Chang Pei and Min-Hung Yeh.
\newblock Two dimensional discrete fractional fourier transform.
\newblock {\em Signal Processing}, 67(1):99--108, 1998.

\bibitem{freeman1991design}
William~T Freeman, Edward~H Adelson, et~al.
\newblock The design and use of steerable filters.
\newblock {\em IEEE Transactions on Pattern analysis and machine intelligence},
  13(9):891--906, 1991.

\bibitem{krizhevsky2012imagenet}
Alex Krizhevsky, Ilya Sutskever, and Geoffrey~E Hinton.
\newblock Imagenet classification with deep convolutional neural networks.
\newblock {\em Advances in neural information processing systems},
  25:1097--1105, 2012.

\bibitem{laptev2016ti}
Dmitry Laptev, Nikolay Savinov, Joachim~M Buhmann, and Marc Pollefeys.
\newblock Ti-pooling: transformation-invariant pooling for feature learning in
  convolutional neural networks.
\newblock In {\em Proceedings of the IEEE conference on computer vision and
  pattern recognition}, pages 289--297, 2016.

\bibitem{esteves2017polar}
Carlos Esteves, Christine Allen-Blanchette, Xiaowei Zhou, and Kostas
  Daniilidis.
\newblock Polar transformer networks.
\newblock {\em arXiv preprint arXiv:1709.01889}, 2017.

\bibitem{sohn2012learning}
Kihyuk Sohn and Honglak Lee.
\newblock Learning invariant representations with local transformations.
\newblock {\em arXiv preprint arXiv:1206.6418}, 2012.

\bibitem{he2015delving}
Kaiming He, Xiangyu Zhang, Shaoqing Ren, and Jian Sun.
\newblock Delving deep into rectifiers: Surpassing human-level performance on
  imagenet classification.
\newblock In {\em Proceedings of the IEEE international conference on computer
  vision}, pages 1026--1034, 2015.

\bibitem{Zhou2017Oriented}
Yanzhao Zhou, Qixiang Ye, Qiang Qiu, and Jianbin Jiao.
\newblock Oriented response networks.
\newblock In {\em Proceedings of the IEEE Conference on Computer Vision and
  Pattern Recognition (CVPR)}, July 2017.

\bibitem{Marcos2017Rotation}
Diego Marcos, Michele Volpi, Nikos Komodakis, and Devis Tuia.
\newblock Rotation equivariant vector field networks.
\newblock In {\em Proceedings of the IEEE International Conference on Computer
  Vision (ICCV)}, Oct 2017.

\bibitem{Worrall2017Harmonic}
Daniel~E. Worrall, Stephan~J. Garbin, Daniyar Turmukhambetov, and Gabriel~J.
  Brostow.
\newblock Harmonic networks: Deep translation and rotation equivariance.
\newblock In {\em Proceedings of the IEEE Conference on Computer Vision and
  Pattern Recognition (CVPR)}, July 2017.

\bibitem{kondor2018generalization}
Risi Kondor and Shubhendu Trivedi.
\newblock On the generalization of equivariance and convolution in neural
  networks to the action of compact groups.
\newblock In {\em International Conference on Machine Learning}, pages
  2747--2755. PMLR, 2018.

\bibitem{cohen2019general}
Taco~S Cohen, Mario Geiger, and Maurice Weiler.
\newblock A general theory of equivariant cnns on homogeneous spaces.
\newblock {\em Advances in neural information processing systems}, 32, 2019.

\bibitem{shen2021pdo}
Zhengyang Shen, Tiancheng Shen, Zhouchen Lin, and Jinwen Ma.
\newblock Pdo-es 2 cnns: Partial differential operator based equivariant
  spherical cnns.
\newblock AAAI 2021.

\bibitem{xie2020color}
Qi~Xie, Qian Zhao, Zongben Xu, and DeYu Meng.
\newblock Color and direction-invariant nonlocal self-similarity prior and its
  application to color image denoising.
\newblock {\em Science China Information Sciences}, 63(12):1--17, 2020.

\bibitem{Brigham1988Fast}
E.~Oran Brigham.
\newblock The fast fourier transform and its applications.
\newblock {\em Prentice-Hall, Inc.}, 1988.

\bibitem{pmlr-v9-glorot10a}
Xavier Glorot and Yoshua Bengio.
\newblock Understanding the difficulty of training deep feedforward neural
  networks.
\newblock In {\em Proceedings of the Thirteenth International Conference on
  Artificial Intelligence and Statistics}, pages 249--256, 2010.

\bibitem{He_2015_ICCV}
Kaiming He, Xiangyu Zhang, Shaoqing Ren, and Jian Sun.
\newblock Delving deep into rectifiers: Surpassing human-level performance on
  imagenet classification.
\newblock In {\em Proceedings of the IEEE International Conference on Computer
  Vision (ICCV)}, December 2015.

\bibitem{Boyd2004Convex}
Boyd Stephen and Vandenberghe Lieven.
\newblock Convex optimization.
\newblock {\em Cambridge university press}, 2004.

\bibitem{Shyh1999Morlet}
Shyh-Jier Huang, Cheng-Tao Hsieh, and Ching-Lien Huang.
\newblock Application of morlet wavelets to supervise power system
  disturbances.
\newblock {\em IEEE Transactions on Power Delivery}, 14(1):235--243, 1999.

\bibitem{Han2013Interpolation}
Dianyuan Han.
\newblock Comparison of commonly used image interpolation methods.
\newblock In {\em Proceedings of the 2nd International Conference on Computer
  Science and Electronics Engineering (ICCSEE 2013)}, pages 1556--1559,
  2013/03.

\bibitem{timofte2017ntire}
Radu Timofte, Eirikur Agustsson, Luc Van~Gool, Ming-Hsuan Yang, and Lei Zhang.
\newblock Ntire 2017 challenge on single image super-resolution: Methods and
  results.
\newblock In {\em Proceedings of the IEEE conference on computer vision and
  pattern recognition workshops}, pages 114--125, 2017.

\bibitem{Larochelle2007Empirical}
Hugo Larochelle, Dumitru Erhan, Aaron Courville, James Bergstra, and Yoshua
  Bengio.
\newblock An empirical evaluation of deep architectures on problems with many
  factors of variation.
\newblock In {\em Proceedings of the 24th International Conference on Machine
  Learning}, page 473–480, 2007.

\bibitem{Diederik2015Adam}
Jimmy~Ba Diederik~Kingma.
\newblock Adam: A method for stochastic optimization.
\newblock In {\em ICLR}, 2015.

\bibitem{huang2015single}
Jia-Bin Huang, Abhishek Singh, and Narendra Ahuja.
\newblock Single image super-resolution from transformed self-exemplars.
\newblock In {\em Proceedings of the IEEE conference on computer vision and
  pattern recognition}, pages 5197--5206, 2015.

\bibitem{martin2001database}
David Martin, Charless Fowlkes, Doron Tal, and Jitendra Malik.
\newblock A database of human segmented natural images and its application to
  evaluating segmentation algorithms and measuring ecological statistics.
\newblock In {\em Proceedings Eighth IEEE International Conference on Computer
  Vision. ICCV 2001}, volume~2, pages 416--423. IEEE, 2001.

\bibitem{zeyde2010single}
Roman Zeyde, Michael Elad, and Matan Protter.
\newblock On single image scale-up using sparse-representations.
\newblock In {\em International conference on curves and surfaces}, pages
  711--730. Springer, 2010.

\bibitem{bevilacqua2012low}
Marco Bevilacqua, Aline Roumy, Christine Guillemot, and Marie~Line
  Alberi-Morel.
\newblock Low-complexity single-image super-resolution based on nonnegative
  neighbor embedding.
\newblock 2012.

\bibitem{lim2017enhanced}
Bee Lim, Sanghyun Son, Heewon Kim, Seungjun Nah, and Kyoung Mu~Lee.
\newblock Enhanced deep residual networks for single image super-resolution.
\newblock In {\em Proceedings of the IEEE conference on computer vision and
  pattern recognition workshops}, pages 136--144, 2017.

\bibitem{zhang2018residual}
Yulun Zhang, Yapeng Tian, Yu~Kong, Bineng Zhong, and Yun Fu.
\newblock Residual dense network for image super-resolution.
\newblock In {\em Proceedings of the IEEE conference on computer vision and
  pattern recognition}, pages 2472--2481, 2018.

\bibitem{zhang2018image}
Yulun Zhang, Kunpeng Li, Kai Li, Lichen Wang, Bineng Zhong, and Yun Fu.
\newblock Image super-resolution using very deep residual channel attention
  networks.
\newblock In {\em Proceedings of the European conference on computer vision
  (ECCV)}, pages 286--301, 2018.

\bibitem{wang2020deep}
Zhihao Wang, Jian Chen, and Steven~CH Hoi.
\newblock Deep learning for image super-resolution: A survey.
\newblock {\em IEEE transactions on pattern analysis and machine intelligence},
  2020.

\bibitem{dong2014learning}
Chao Dong, Chen~Change Loy, Kaiming He, and Xiaoou Tang.
\newblock Learning a deep convolutional network for image super-resolution.
\newblock In {\em European conference on computer vision}, pages 184--199.
  Springer, 2014.

\bibitem{he2016deep}
Kaiming He, Xiangyu Zhang, Shaoqing Ren, and Jian Sun.
\newblock Deep residual learning for image recognition.
\newblock In {\em Proceedings of the IEEE conference on computer vision and
  pattern recognition}, pages 770--778, 2016.

\bibitem{huang2017densely}
Gao Huang, Zhuang Liu, Laurens Van Der~Maaten, and Kilian~Q Weinberger.
\newblock Densely connected convolutional networks.
\newblock In {\em Proceedings of the IEEE conference on computer vision and
  pattern recognition}, pages 4700--4708, 2017.

\bibitem{wang2004image}
Zhou Wang, Alan~C Bovik, Hamid~R Sheikh, and Eero~P Simoncelli.
\newblock Image quality assessment: from error visibility to structural
  similarity.
\newblock {\em IEEE transactions on image processing}, 13(4):600--612, 2004.

\bibitem{bracewell1983discrete}
{Ronald~N Bracewell.
\newblock Discrete Hartley Transform.
\newblock {\em JOSA}, 73(12): 1832--1835, 1983.}

\bibitem{ahmed1974discrete}
{Nasir Ahmed, T\_ Natarajan, and Kamisetty~R Rao.
\newblock Discrete Cosine Transform.
\newblock {\em IEEE transactions on Computers}, 100(1):90--93, 1974.}

\end{thebibliography}

%\vspace{-9mm}
\begin{IEEEbiography}[{\includegraphics[width=1in,height=1.25in,clip,keepaspectratio]{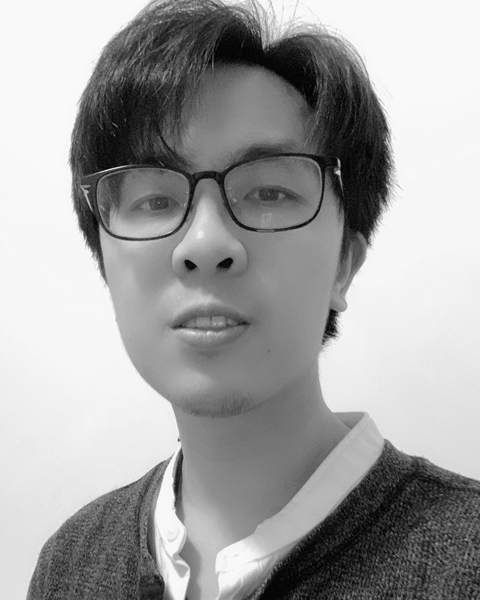}}]{Qi Xie} received the B.Sc. and Ph.D degree from Xi'an Jiaotong University, Xi'an, China, in 2013 and 2020 respectively. He was a Visiting Scholar with Princeton University, Princeton, NJ, USA, from 2018 to 2019.
He is currently an assistant professor with School of Mathematics and Statistics, Xi'an Jiaotong University.
His current research interests include model-based deep learning and filter parametrization-based deep learning.
%\vspace{-0mm}
\end{IEEEbiography}
%\vspace{-60mm}
\begin{IEEEbiography}[{\includegraphics[width=1in,height=1.25in,clip,keepaspectratio]{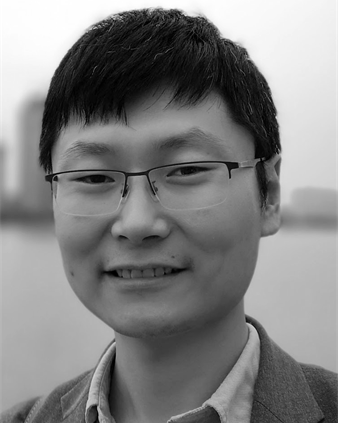}}]{Qian Zhao}  received the B.Sc. and Ph.D degrees from Xi'an Jiaotong University, Xi'an, China, in 2009 and 2015, respectively.
He was a Visiting Scholar with Carnegie Mellon University, Pittsburgh, PA, USA, from 2013 to 2014. He is currently an associate professor with School of Mathematics and Statistics, Xi'an Jiaotong University. His current research interests include low-rank matrix/tensor analysis, Bayesian modeling and meta learning.
%\vspace{9mm}
\end{IEEEbiography}
%\vspace{-60mm}
\begin{IEEEbiography}[{\includegraphics[width=1in,height=1.25in,clip,keepaspectratio]{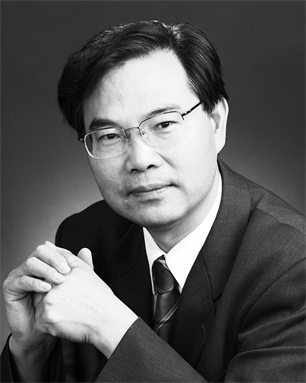}}]{Zongben Xu} received the Ph.D. degree in mathematics from Xi'an Jiaotong University, Xi'an, China, in 1987. He currently serves as the Academician of the Chinese Academy of Sciences, the Chief Scientist of the National Basic Research Program of China (973 Project), and the Director of the Institute for Information and System Sciences with Xi'an Jiaotong University. His current research interests include nonlinear functional analysis and intelligent information processing.
He was a recipient of the National Natural Science Award of China in 2007 and the winner of the CSIAM Su Buchin Applied Mathematics Prize in 2008.
\end{IEEEbiography}
%\vspace{-60mm}
\begin{IEEEbiography}[{\includegraphics[width=1in,height=1.25in,clip,keepaspectratio]{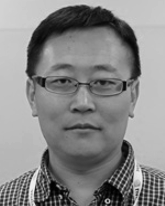}}]{Deyu Meng} received the B.Sc., M.Sc., and Ph.D. degrees from Xi'an Jiaotong University, Xi'an, China, in 2001, 2004, and 2008, respectively. He is currently a professor with School of Mathematics and Statistics, Xi'an Jiaotong University, and adjunct professor with Faculty of Information Technology, The Macau University of Science and Technology. From 2012 to 2014, he took his two-year sabbatical leave in Carnegie Mellon University. His current research interests include model-based deep learning, variational networks, and meta-learning.
%\vspace{-0mm}
\end{IEEEbiography}

\end{document}